%% file: Manuscript.tex
\documentclass[preprint,12pt]{elsarticle}

\usepackage{epsr}
\usepackage{amssymb}
\usepackage{amsmath}

\input mymath

\usepackage[linesnumbered,ruled,vlined, ruled,vlined]{algorithm2e}
\usepackage{amsmath}
\usepackage{lineno}
\usepackage{pgfplots}
\pgfplotsset{compat=1.18}
\usepackage{subcaption}
\usepackage{tikz}
\usepackage{xcolor}
\usepgfplotslibrary{fillbetween}
\usepackage{float}

\def\eg{e.g., }
\def\ie{i.e., }
\def\etal{et al. }

\title{On Approximating the Dynamic Response of Synchronous Generators via Operator Learning: A Step Towards Building Deep Operator-based Power Grid Simulators}

\journal{Electric Power Systems Research}
\begin{document}
\begin{frontmatter}


\author[label2]{Christian~Moya}
\author[label3]{Amirhossein Mollaali}
\author[label2,label3]{Guang Lin\corref{core}}
	\ead{guanglin@purdue.edu}
    \cortext[core]{corresponding author}
\author[label4]{Meng~Yue}
 \affiliation[label2]{organization={Purdue University},
			 addressline={Department of Mathematics},
		 	 city={West Lafayette},
	 		 postcode={47906},
			 state={IN},
			 country={USA}}
 \affiliation[label3]{organization={Purdue University},
             addressline={School of Mechanical Engineering},
             city={West Lafayette},
             postcode={47906},
             state={IN},
             country={USA}}
 \affiliation[label4]{organization={Brookhaven National Laboratory},
             addressline={Interdisciplinary Science Department},
             city={Upton},
             postcode={11973},
             state={NY},
             country={USA}}
             
\begin{abstract}
This paper develops an Operator Learning framework for approximating the dynamic response of synchronous generators. The framework can be used to (i) build a neural network-based generator model that interacts with a power grid simulator or (ii) shadow the true generator's transient response. First, we develop a data-driven Deep Operator Network (DeepONet) to approximate the infinite-dimensional solution operator of the generators. Then, we design a numerical scheme based on DeepONet that simulates the generator's response over a given time horizon. The proposed scheme recursively employs the trained DeepONet to simulate the response for a given multi-dimensional input that describes the interaction between the generator and the power grid. In addition, we design a residual DeepONet numerical scheme that can incorporate information from existing mathematical models. We accompany this residual DeepONet scheme with an estimate for the prediction's cumulative error. Finally, we build a data aggregation (DAgger) strategy that allows fine-tuning of DeepONets using aggregated training data that the DeepONets will likely encounter during interactive simulations with other grid components. As a proof of concept, we demonstrate that the proposed frameworks can effectively approximate the transient model of a synchronous generator.
\end{abstract}

\end{frontmatter}



\section{Introduction} \label{sec:introduction}
\input{1-introduction}
\section{Problem Formulation} 
\label{sec:problem-formulation}
\input{2-problem-formulation}
\section{The Deep Operator Learning Framework} \label{sec:DeepONet}
\input{3-operator-learning}
\section{Incorporating Mathematical Models} \label{sec:residual-DeepONet}
\input{4-residual-learning}
\section{The Data Aggregation~(DAgger) Algorithm} \label{sec:DAgger}
\input{5-DAgger}
\section{Numerical Experiments} \label{sec:numerical-experiments}
\input{6-numerical-experiments}

\section{Discussion and Future Work} \label{sec:discussion}
Experimental results in the previous section demonstrate that the DeepONet framework can effectively simulate the dynamic response of power grid components interacting with the power grid. Let us now give a preamble of our future work.

\textit{On moving towards digital twins.} We will develop a data assimilation method to calibrate DeepONet online. This will allow a \textit{virtual} replica based on DeepONet to shadow the physical components of the power grid.

\textit{On securing the privacy of the power grid component's owners.} This paper assumes that we can collect training data from multiple generator owners, but in reality, they may want to protect their privacy. Thus, our future work will develop federated learning protocols for DeepONet training~\cite{moya2022fed}, which will secure proprietary models and data privacy.

\textit{On predicting large-scale power grids.} Our goal is to use the proposed DeepONet framework in a composable manner to construct large-scale power grid simulators and digital twins. These composable large-scale simulators will then be used to optimize, control, and quantify the uncertainty of smart grids with high penetration of renewable energy sources.
\section{Conclusion} \label{sec:conclusion}
This paper proposed a Deep Operator Network (DeepONet) framework, which can learn the dynamic response of synchronous generators (SG) that interact with a simulator or power grid. The DeepONet approximates the SGs' solution operator for any time interval and predicts the dynamic response for a given time horizon. We also developed and estimated the cumulative error of a residual DeepONet that incorporates information from mathematical models. For SGs interacting with a simulator, we designed a data aggregation (DAgger) algorithm that trains and fine-tunes DeepONets using collected data inputs that the DeepONet will likely encounter when interacting with the simulator. Finally, we verified the efficacy of the proposed frameworks by learning an SG that interacts with an infinite bus.

\appendix
\section{Shadowing a Transient Synchronous Generator} \label{appendix:pst-experiment}
This experiment uses DeepONet to approximate the response of an SG using data collected with the Power System Toolbox (PST)~\cite{chow1992toolbox}. The focus is on the PST transient model of a generator with a default exciter interacting with the classical two-area system. Unlike the numerical experiments in Section~\ref{sec:numerical-experiments}, this experiment only "shadows" the SG response using the trained DeepONet. Thus, PST does not use the DeepONet's predicted state $x(t_n +h)$ to solve~\eqref{eq:multimachine-stator} and~\eqref{eq:multimachine-network}, and the DeepONet always observes the correct interface inputs~$y$, simplifying the learning task and reducing error accumulation. With this scenario, we aim to represent the problem of learning the response of an SG connected to an actual power grid. This is the first step towards building an SG digital twin.

\textit{Training data.} We generated training data $\mathcal{D}_{\text{PST}}$ by simulating $N_\text{exp}=300$ experiments on PST. Each experiment consisted of simulating a two-area system using a uniform partition $\mathcal{P} \subset [0.0,5.0]$ (s), with a constant step size of $h = 0.05$. A fault was simulated at time $t_f = 0.1 (s)$ and cleared at time $t_f + \Delta t_f$, where $\Delta t_f$ is a random fault duration sampled from the interval $[0.01, 0.1]$ (s). Trajectory data was collected after each experiment, including the interacting input trajectories $\{y(t_n):t_n \in \mathcal{P}\}$, the exciter input data $\{u(t_n)\equiv E_\text{fld}(t_n):t_n \in \mathcal{P}\}$, and state trajectory data $\{x(t_n):t_n \in \mathcal{P}\}$.
We constructed our training dataset using this trajectory data. In particular, we discretized the inputs using interpolation and $m=2$ sensors, \ie $\tilde{y}^n_m :=\{y(t_n+d_0), y(t_n + d_1))\}$ and $\tilde{u}^n_m :=\{u(t_n+d_0), u(t_n + d_1))\},$ where $d_0=0.0$ and $d_1$ was uniformly sampled from the open interval $(0,h)$. The final training dataset is:
$\mathcal{D}_\text{PST} = \{x_k(t_n),\tilde{y}^n{1,k}, \tilde{u}^n{m,k}, \{0,d_{m,k}\}, h_k, G_{\Delta,k}\}{k=1}^{N\text{train}}.$

\begin{figure}[t!]
\centering
\begin{subfigure}[b]{0.4\textwidth}
\centering
\includegraphics[width=0.99\textwidth]{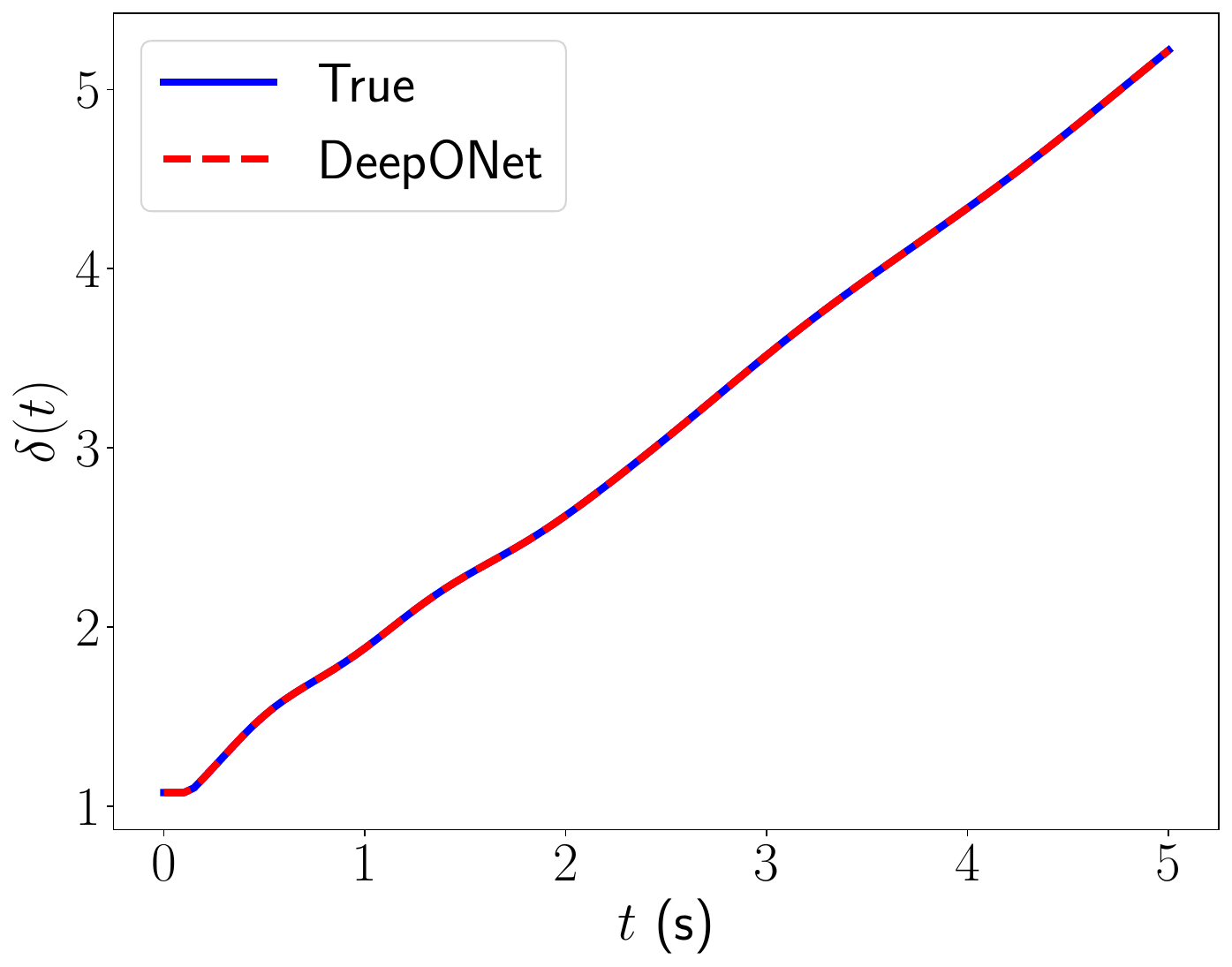}
\end{subfigure}
\begin{subfigure}[b]{0.4\textwidth}
\centering
\includegraphics[width=0.99\textwidth]{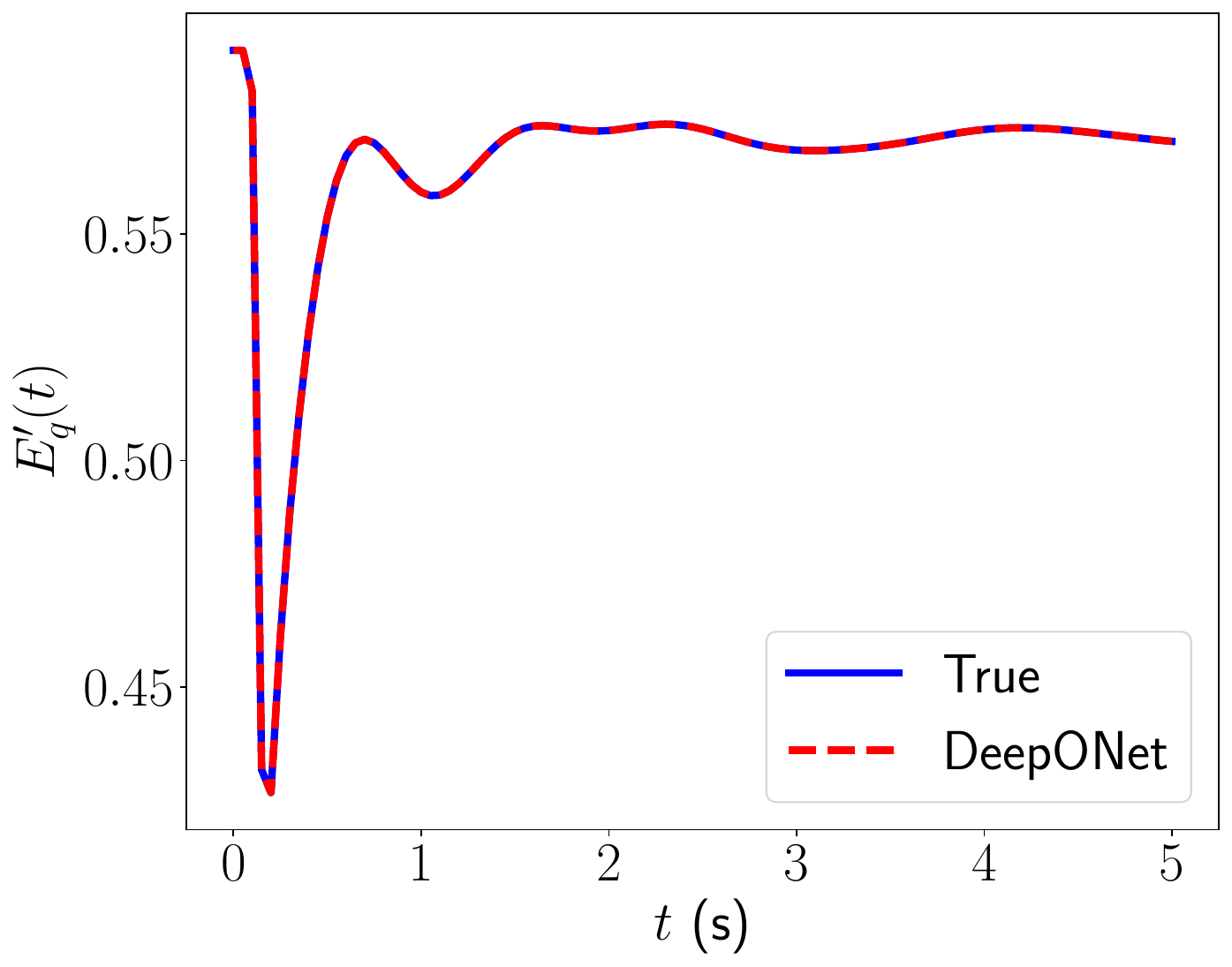}
\end{subfigure}
\caption{Comparison of the data-driven DeepONet prediction with the true fault test trajectory of the SG selected states $ (\delta(t), E_q'(t))^\top$. We simulated the test trajectory using the Power System Toolbox~(PST) over the uniform partition $\mathcal{P} \subset [0,5]$ (s) of constant step size $h=0.05$.}
\vspace{-1.0em}
\label{fig:pst-results}
\end{figure}

\textit{Training protocol and test results.} We used Adam to train the DeepONet. We optimized the Branch and Trunk nets' architectures with a simple hyper-parameter routine. Finally, we tested DeepONet using a PST test trajectory not included in the training dataset. Fig.~\ref{fig:pst-results} shows that DeepONet accurately predicted the dynamic response of the generator.

\section*{Acknowledgement}
GL, AM and CM gratefully acknowledge the support of Brookhaven National Laboratory Subcontract 382247, National Science Foundation (DMS-2053746, DMS-2134209, ECCS-2328241, CBET-2347401 and OAC-2311848), and U.S.~Department of Energy (DOE) Office of Science Advanced Scientific Computing Research program DE-SC0023161, and DOE–Fusion Energy Science, under grant number: DE-SC0024583. 

\section*{Data and code availability}

The data and code that support the findings of this study are available from the corresponding author upon reasonable request, subject to institutional and project-related restrictions.


\bibliographystyle{elsarticle-num} 
\bibliography{references.bib}
\end{document}

%% file: mymath.tex
\usepackage[linesnumbered,ruled,vlined, ruled,vlined]{algorithm2e}
\usepackage{amsmath}
\usepackage{amssymb}
\usepackage{amsthm}

\theoremstyle{remark}
\newtheorem*{remark}{Remark}
\theoremstyle{definition}
\newtheorem{theorem}{Theorem}[section]

\newtheorem{lemma}[theorem]{Lemma}

\usepackage{graphicx}
\usepackage{caption}
\usepackage{subcaption}
\usepackage{tikz}

\def\eg{\emph{e.g., }}
\def\ie{\emph{i.e., }}
\def\etal{\emph{et al. }}

%% file: 1-introduction.tex
The smart grid~\cite{arnold2011challenges} promises to improve significantly the reliability and cost-effectiveness of electricity production and delivery. To fulfill this promise, the smart grid must enable multiple innovative technologies, such as a growing fleet of electric vehicles, a vast number of distributed renewable energy resources, and an increasing number of responsive loads. As a result, the power systems community must design and develop new, state-of-the-art methods that will support the operation, prediction, and control of the future smart grid~\cite{henriquez2021transient}.

For decades, the power systems community has developed sophisticated numerical simulation schemes~\cite{milano2010power,sauer2017power} to predict the dynamic response of the power grid after a disturbance event~\cite{milano2010power}. However, the high computational cost of these schemes has prompted power utilities to recognize the need for faster simulation tools~\cite{flueck2014high}. Such tools are necessary to optimize and control the future smart grid. Unfortunately, the current state-of-the-art simulation schemes may be prohibitively expensive for the often complex forward simulations required for such control and optimization tasks.
\vspace{-0.5em}
\subsection{Prior Works} \label{sub-sec:prior-works}
\textit{Power grid time-domain simulation.} Most commercial power grid simulators use either a partitioned-solution approach or an implicit integration scheme \cite{milano2010power}. The partition approach solves power grid DAEs sequentially, freezing the dynamic (resp. algebraic) variables to solve for the algebraic (resp. dynamic) variables. This strategy allows for independent simulation of power grid dynamic components but introduces a delay error that propagates throughout the simulation \cite{stott1979power}. On the other hand, implicit integration schemes simultaneously solve the power grid DAEs, eliminating delay error propagation. However, they require solving all power grid dynamic components simultaneously using expensive iterative methods (\eg Newton’s method) \cite{milano2010power}.

\textit{Parallel computing-based solutions.} Advancements in high-performance parallel computing have led to the development of solutions to speed up the simulation of power grids~\cite{tomim2009parallel, palmer2016gridpacktm}. However, using parallel computing requires a large amount of computational resources, and the speedup is limited by the serial nature of the power grid applications. This limitation could prevent their use in tasks that require multiple forward simulations, such as control, or optimization.

\textit{Deep learning-based solutions.} Deep learning-based computational tools have the potential to revolutionize how we predict and simulate complex dynamical systems. Recently, numerous publications have developed faster alternatives to traditional numerical schemes. These publications can be roughly classified into deep learning-based methods aiming at (i) learning the governing equations of the system~\cite{brunton2016discovering,schaeffer2017learning}, (ii) encoding the underlying physics into training protocols~\cite{raissi2019physics, moya2022dae, karniadakis2021physics}, or (iii) predicting the future states of the system~\cite{qin2019data, jin2022learning}.

For example, Brunton \etal~\cite{brunton2016discovering} learned the unknown governing equations of a dynamic system using data. Similarly, the authors in~\cite{schaeffer2017learning} used data to learn a sparse representation of partial differential equations. In~\cite{misyris2020physics}, the authors applied the physics-informed neural network framework introduced in~\cite{raissi2019physics, karniadakis2021physics} to encode a simplified version of the power grid's dynamics into the training loss function. In our previous work~\cite{moya2022dae}, we designed a physics-informed neural network that incorporates differential-algebraic equations into training protocols, which often describe the power grid dynamics.

On the other hand, several works~\cite{qin2019data,jin2022learning} propose learning the system's next-step response and recursively using it to predict the system's evolution for given time horizons. One can implement such a next-step strategy using residual networks~(resNets)~\cite{qin2019data}. However, most of the deep learning-based methods mentioned above avoid overfitting by using large amounts of data, and they may need to be retrained when dealing with unseen operating conditions. Therefore, the effective design of deep learning-based solutions for predicting and simulating large-scale, complex dynamical systems remains an open research question.

\textit{Component-based learning.} To enable effective learning strategies for large-scale dynamical systems, one can follow a component-based design approach. This approach involves learning the input/output response of individual components. By synchronizing these trained components, we can then predict and simulate large-scale dynamical systems. In \cite{li2020machine}, we introduced this component-based learning strategy to the power systems community. We used recurrent neural networks to approximate the input/output response of a synchronous generator (SG) interacting with the rest of the grid. However, this strategy required (i) a large amount of data to control error propagation and (ii) a uniform simulation step size.

In a related effort, the authors of~\cite{anantharaman2021composable} proposed using continuous-time echo state networks to build a faster surrogate neural architecture. This architecture can approximate the input/output response of a library of components and discretely interact with causal models through functional mockup interfaces~\cite{blochwitz2011functional}. However, the proposed architecture ignores the infinite-dimensional nature of the problem, lacks the mathematical foundation of the original echo state networks~\cite{ozturk2007analysis}, and requires rolling a numerical integration scheme to obtain the current prediction. To address the aforementioned problems, this paper proposes using the mathematically sound infinite-dimensional framework of operator learning, which we introduced to the power systems community in~\cite{moya2022deeponet}.

\textit{Operator learning.} In their seminal paper~\cite{lu2021learning}, Lu~\etal designed a Deep Operator Network~(DeepONet) framework based on the universal approximation theorem for nonlinear operators~\cite{chen1995universal}. This framework can learn nonlinear operators, which are mappings between infinite-dimensional spaces. Compared to traditional neural networks, DeepONet learns with streams of scattered data and exhibits minor generalization errors. The effectiveness of DeepONet has been demonstrated in various applications, such as control systems~\cite{li2022learning}, power systems~\cite{moya2022deeponet}, and multi-physics problems~\cite{cai2021deepm}. Extensions to the original DeepONet framework have enabled dealing with stiff dynamics~\cite{wang2021long} and discontinuities~\cite{lanthaler2022error}, incorporating physics-informed training~\cite{wang2021long}, and quantifying uncertainty~\cite{moya2022deeponet,lin2021accelerated,yang2022scalable}. Thus, in this paper, we adopt an operator learning framework to approximate the solution operator of power grid dynamic components, such as synchronous generators (SGs), that interact with the rest of the power grid or a numerical simulator.
\vspace{-0.5em}
\subsection{Our Work} \label{sub-sec:our-work}
The objectives of this paper are described next.
\begin{enumerate}
    \item \textit{Learning the solution operator.} Our goal is to create an operator learning framework that can approximate the dynamic response of a library of grid components (such as SGs) as they interact with the rest of a power grid over any given local time interval.
    \item \textit{Short- and medium-term simulation.} We then aim to design a recursive numerical scheme that uses the proposed framework to predict the dynamic response of SGs to a range of operating conditions.
    \item \textit{Incorporating power grid's mathematical models.} Our final goal is to incorporate the mathematical models developed by the power community into our framework.
\end{enumerate}
We detail the contributions of this paper next.
\begin{itemize}
    \item We design (in Section~\ref{sec:DeepONet}) a Deep Operator Network (DeepONet) that approximates the local solution operator of an SG. The DeepONet uses three inputs: (i) the current state, (ii) the current interface input that describes the interaction of the SG with the rest of the power grid, and (iii) an arbitrary prediction timestep. DeepONet then outputs the desired future state over an arbitrary time interval. Our framework is the first to use operator learning to approximate the solution operator for dynamic components of the power grid.
    \item We then (in Section~\ref{sub-sec:DeepONet-scheme}) develop a numerical scheme that recursively simulates the generator's response over a short/medium-term horizon using the trained DeepONet. 
    \item We also propose (in Section~\ref{sec:residual-DeepONet}) a residual DeepONet that learns the residual dynamics obtained by comparing the actual solution operator with the solution operator obtained from (i) a mathematical model of the SG or (ii) a physics-informed neural network. 
    \item We theoretically estimate (in Section~\ref{sub-sec:error-bound}) the cumulative error of the proposed residual DeepONet-based numerical scheme.
    \item Finally, for an SG interacting with a simulator, we propose (in Section~\ref{sec:DAgger}) a data aggregation (DAgger) algorithm. This algorithm enables (i) supervised training of DeepONets with small datasets and (ii) fine-tuning of DeepONets by aggregating datasets that the DeepONet will likely encounter during interaction with the simulator.
\end{itemize}
We remark that the framework we will develop for SGs can also be applied to other dynamic components of the power grid, including inverter-based generators.

We organize the rest of this paper as follows. Section~\ref{sec:problem-formulation} describes how to approximate the solution operator of an SG interacting with a power grid. Then, Section~\ref{sec:DeepONet} presents a data-driven DeepONet that learns this solution operator locally. Using the trained DeepONet, we design a numerical scheme that simulates the SG's response over a given time horizon in Section~\ref{sub-sec:DeepONet-scheme}. In Section~\ref{sec:residual-DeepONet}, we design and estimate the cumulative error of a residual DeepONet that learns to use available mathematical models of SGs. Section~\ref{sec:DAgger} designs a data aggregation (DAgger) for training and fine-tuning DeepONets. As a proof of concept, we illustrate the effectiveness of the proposed frameworks by learning a transient model of an SG in Section~\ref{sec:numerical-experiments}. Finally, Section~\ref{sec:discussion} discusses our results and future work, and  Section~\ref{sec:conclusion} concludes the paper.

%% file: 2-problem-formulation.tex
This paper aims to approximate the dynamic response of a synchronous generator (SG) interacting with a power grid or a numerical simulator such as the Power Systems Toolbox (PST)~\cite{chow1992toolbox}. We model the SG using the following non-autonomous initial value problem (IVP):
\begin{gather}
\begin{aligned} \label{eq:nonautonomous-system}
    \frac{d}{dt}x(t) = f(x(t),y(t), u(t);\lambda),~x(0) = x_0,~t \in [0,T].
\end{aligned}    
\end{gather}
Here, $x(t) \in \mathcal{X} \subseteq \mathbb{R}^{n_x}$ is the vector-valued state function, $y(t) \in \mathcal{Y} \subseteq \mathbb{R}^{n_y}$ are the interacting variables, $u(t) \in \mathcal{U} \subseteq \mathbb{R}^{n_u}$ is the control input, $\lambda \in \Lambda \subset \mathbb{R}^{n_\lambda}$ are the parameters of a specific SG, $[0, T] \subset [0, + \infty)$ is the finite time horizon of interest, and $x_0 \in \mathcal{X}$ is the initial condition. We will assume that the vector field $f: \mathcal{X} \times \mathcal{Y} \times \mathcal{U} \to \mathcal{X}$ is (i) unknown (in Section~\ref{sec:DeepONet}) and (ii) partially known (in Section~\ref{sec:residual-DeepONet}), \ie we have some approximate model such that $f_\text{approx} \approx f$.

\begin{figure}[t!]
\centering
\includegraphics[width=.55\textwidth]{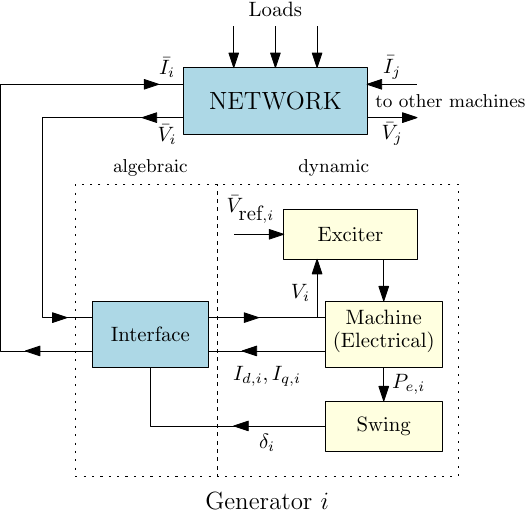}
\caption{A \textit{block diagram} of the differential-algebraic model~\eqref{eq:multimachine-dae} introduced in~\cite{sauer2017power}. The model describes the Network equations~\eqref{eq:multimachine-network}, the stator (\ie generator interface) algebraic equations~\eqref{eq:multimachine-stator}, and the SGs' differential equations~\eqref{eq:multimachine-gens}.}
\label{fig:multimachine-block}
\vspace{-1.5em}
\end{figure}

One can use the IVP~\eqref{eq:nonautonomous-system} to (i) describe the dynamic response of an SG interacting with a numerical simulator, \eg PST (see Fig.~\ref{fig:multimachine-block}) or (ii) shadow the dynamic response of an actual generator. More specifically, we consider a multi-machine power grid described using the following differential-algebraic~(DA) model introduced in Sauer~\etal \cite{sauer2017power}: 
\begin{subequations} \label{eq:multimachine-dae}
\begin{align}
\dot{z} &= f_o(z, I_{d-q}, \bar{V}, u) \label{eq:multimachine-gens} \\
I_{d-q} &= h(z, \bar{V}) \label{eq:multimachine-stator} \\
0 &= g(z, I_{d-q}, \bar{V}). \label{eq:multimachine-network}
\end{align}
\end{subequations}
In the above, \eqref{eq:multimachine-gens} describes the SGs' dynamic equations, \eqref{eq:multimachine-stator} the stator algebraic equations, and \eqref{eq:multimachine-network} the network equations. Within this DA model, the proposed operator learning framework can provide the dynamic response of a \textit{parametric family} of SGs, \ie $f(\cdot;\lambda_i) \approx f_{o,i}$ for some $i$'s. In this case, the interacting variables are (see Fig.~\ref{fig:multimachine-block}) the generator's stator currents and bus terminal voltage, \ie $y(t) = (I_d(t), I_q(t), V(t))^\top$, and the control input is the voltage regulator field voltage, \ie $u(t) \equiv E_{fd}(t)$. However, we remark that the proposed method can use other interacting variables, such as bus active and reactive powers, which we used in our previous work~\cite{li2020machine}. 

\emph{Focusing on interacting variables.} This paper solely focuses on the interacting variables~$y(t)$, while neglecting the control input~$u(t)$, and fixing the parameters~$\lambda$. This is because the analysis and learning protocols for~$u(t)$ and $\lambda$ are similar to those for~$y(t)$.

Furthermore, the interacting variables~$y(t)$ propagate prediction errors when interacting with a simulator that describes the rest of the grid. Thus, it is imperative to design robust deep-learning methods to control this error propagation. In the absence of the control input~$u(t)$, the dynamics of the non-autonomous IVP \eqref{eq:nonautonomous-system} are written as follows:
\begin{gather}
\begin{aligned} \label{eq:nonautonomous-system-no-control}
\frac{d}{dt}x(t) = f(x(t),y(t);\lambda),~x(0) = x_0,~ t \in [0,T].
\end{aligned}
\end{gather}

\textit{The solution operator.} We describe the dynamic response of the selected SG using the \textit{solution operator} of~\eqref{eq:nonautonomous-system-no-control}, denoted as~$G$. $G$ takes the initial condition~$x(0) = x_0 \in \mathcal{X}$ as input, along with a continuous sequence of interacting variables denoted, with some abuse of notation, as $y_{[0,t)}:= \{y(s) \in \mathcal{Y}: s \in [0,t)\}$. $G$ then computes the state $x(t) \in \mathcal{X}$ at time $t \in [0,T]$ as follows:
\begin{align*}
    G\left(x_0, y_{[0,t)};\lambda \right)(t) \equiv x(t) = x_0 + \int_{0}^t f(x(s),y(s);\lambda) ds.
\end{align*}

\textit{Approximate non-autonomous IVP.} In practice, however, we may only have access to a discrete representation of the trajectory of $y(t)$ within the interval~$[0,T]$. We denote such an approximation as~$\tilde{y}$. Using~$\tilde{y}$ in~\eqref{eq:nonautonomous-system-no-control} results in the following approximate non-autonomous IVP:
\begin{gather}
\begin{aligned} \label{eq:nonautonomous-approximate}
    \frac{d}{dt}\tilde{x}(t) &= f(\tilde{x}(t),\tilde{y};\lambda),~\tilde{x}(0) = x_0,~t \in [0,T].
\end{aligned}    
\end{gather}
Its solution operator~$\tilde{G}$ at time $t \in [0,T]$ is
\begin{align*}
    \tilde{G}\left(x_0, \tilde{y};\lambda\right)(t) \equiv \tilde{x}(t) = x_0 + \int_{0}^t f(\tilde{x}(s),\tilde{y};\lambda) ds.
\end{align*}
In Lu~\etal~\cite{lu2021learning}, the authors introduced a Deep Operator Network (DeepONet) to approximate the solution operator~$G$ (with $x_0 = 0$) and demonstrated its approximation capacity via~$\tilde{G}$. The input to the proposed DeepONet was the trajectory of the function~$y_{[0, T]}$, discretized using~$m \ge 1$ interpolation points (also known as \textit{sensors} in~\cite{lu2021learning}). However, their proposed DeepONet~\cite{lu2021learning} has two drawbacks.

(i) Their DeepONet effectively approximates $G$ (with $x_0 = 0$) for small values of~$T$. For longer time horizons, \ie for $T \gg 1$ second, increasing the number~$m$ of sensors becomes necessary, which makes the training process challenging.

(ii) To predict~$x(t)$ for any $t \in [0,T]$, their DeepONet requires the trajectory $u(t)$ within the \textit{entire} interval~$[0,T]$ as input. However, this assumption does not hold for the problem studied in this paper. Here, we only have access to past values of $y(t)$, \ie within the interval $[0,t) \subset [0,T]$. Thus, the DeepONet designed in~\cite{lu2021learning} is not suitable for the problem we are addressing in this paper.

To alleviate the two aforementioned drawbacks, we propose an operator learning framework in the next section. The framework first designs and trains a novel DeepONet to locally approximate $G$, and then recursively uses the trained DeepONet to approximate the SG's dynamic response over the entire interval of $[0, T]$.

%% file: 3-operator-learning.tex
The proposed operator learning framework aims to approximate the \textit{local} solution operator, \ie the solution operator~$G$ within the arbitrary time interval~$[t_n, t_n+h_n] \subset [0,T]$:
\begin{align*}
    \tilde{G}\left(\tilde{x}(t_n), \tilde{y}_m^n, \lambda \right)(h_n) &\equiv \tilde{x}(t_n + h_n) \\
    &= \tilde{x}(t_n) + \int_{0}^{h_n} f(\tilde{x}(t_n+ s),\tilde{y}^n_m;\lambda) ds,
\end{align*}
where $\tilde{y}_m^n$ is the \textit{discretized} input~$y(t)$ within the interval~$[t_n,t_n + h_n]$ using $m \ge 1$ sensors, \ie 
$\tilde{y}_m^n=(y(t_n + d_0), y(t_n+d_1), \ldots, y(t_n + d_{m-1})) \approx y_{[t_n,t_n + h_n]}.$ Here, $h_n \le h$ is the step-size, $h > 0$ the given maximum step-size, and $\{d_{k-1}\}_{k=1}^m \subset [0,h_n]$ the relative sensor locations.
\begin{remark}
\textit{The case of $m=1$ sensors.} Let's consider an SG interacting with a simulator. When the number of sensors is only one, \ie $m=1$, the discretized interacting variable is the singleton $\tilde{y}_m^n \equiv y(t_n)$. This scenario is similar to the integration step of the SG's dynamics for the partition-solution approach~\cite{milano2010power}, which is used in PST~\cite{chow1992toolbox} and extensively implemented in our numerical experiments (Section~\ref{sec:numerical-experiments}).
\end{remark}
\vspace{-0.5em}
\subsection{The Deep Operator Network} \label{sub-sec:DeepONet-design}
\begin{figure}[t!]
\centering
\includegraphics[width=0.85\textwidth]{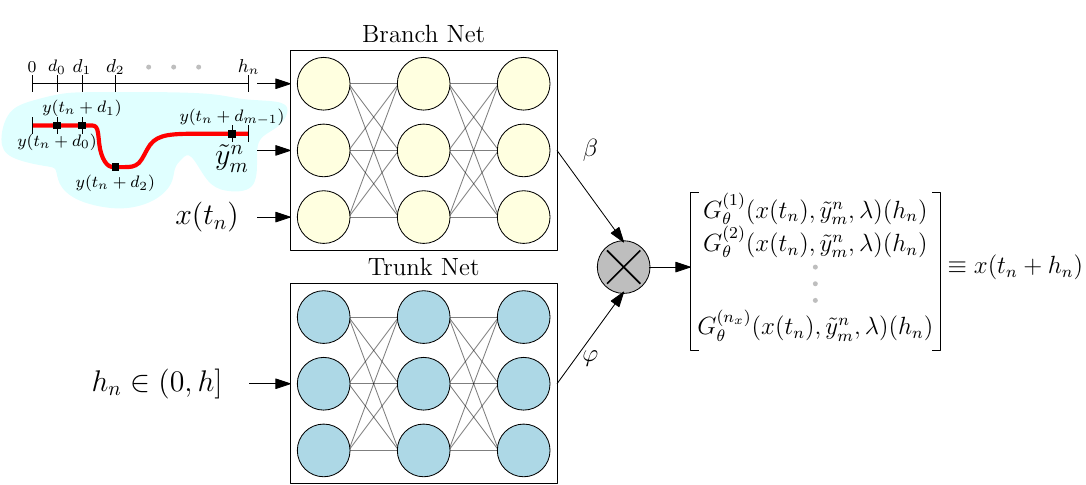}
\caption{DeepONet for learning the solution operator describing the dynamic response of an SG. The Branch Net takes as input a vector resulting from concatenating (i) the current state~$\tilde{x}(t_n)$, (ii) the discretized interacting variables~$\tilde{y}_m^n$, and (iii) the relative/flexible sensor locations $ \{ d_{k-1}\}_{k=1}^m$. Then, the Branch Net outputs the vector of trainable coefficients~$\beta$. The Trunk Net takes as the input $h_n$ and outputs the vector of trainable basis functions~$\varphi$. Finally, we obtain the DeepONet's output using \textit{linear trainable approximations of the states}, \ie a dot product between~$\beta$ and $\varphi$.}
\label{fig:DeepONet}
\vspace{-1.5em}
\end{figure}

To approximate the \textit{local} solution operator, we draw inspiration from successful strategies used in the numerical analysis of partial differential equations~\cite{iserles2009first}. Specifically, we aim to approximate the solution operator using a \textit{trainable} linear representation. In this approach, the coefficients of the trainable linear representation will process the inputs $x(t_n)$ and $\tilde{y}^n_m$, while the corresponding basis functions will process location points $h_n$ within the output function domain $[0,h]$. Such a linear model will lead to a novel multi-output \textit{Deep Operator Network} (DeepONet) $G_\theta$, with a vector of trainable parameters $\theta \in \mathbb{R}^p$. Our DeepONet is composed of \textit{two} neural networks: the \textit{Branch} Net and the \textit{Trunk} Net (see Fig.~\ref{fig:DeepONet}).

The \textit{Branch} Net takes as input (i) the current state~$x(t_n) \in \mathcal{X}$ and (ii) the trajectory of the interacting variables~$\tilde{y}_m^n$ discretized using $m \ge 1$ sensors. The Branch Net then outputs the vector of trainable coefficients $\beta \in \mathbb{R}^{q n_x}$.

To enable the recursive numerical scheme described in Section~\ref{sub-sec:DeepONet-scheme}, we must design the \textit{Branch} Net to be resolution-independent. This means it must allow flexible locations for the $m$ sensors. To achieve this, we let the Branch Net take as input the concatenation of the vector of relative sensor locations, $\{d_{k-1}\}_{k=1}^m \subset [0,h_n]$, with the input values $\tilde{y}_m^n$, as shown in Fig.~\ref{fig:DeepONet}. However, we remark that for the critical case of $m=1$, \ie when we use the partition approach~\cite{milano2010power}, there is no need to include the relative sensor locations.

The \textit{Trunk} Net takes the step-size $h_n \in (0, h_n]$ as input and outputs the vector of trainable basis functions:
$\varphi:= (\varphi_1(h_n), \varphi_2(h_n), \ldots, \varphi_{qn_x}(h_n))^\top \in \mathbb{R}^{q n_x}.$
Finally, we compute each component of the DeepONet's vector output (\ie each predicted state) $G_\theta(x(t_n),\tilde{y}_m^n,\lambda)(h_n) \equiv x(t_n + h_n)$, using a linear representation (\ie via the dot product):
\begin{align*}
    x^{(i)}(t_n + h_n) &\equiv  G^{(i)}_\theta(x(t_n),  \tilde{y}_m^n; \lambda)(h_n) \\ &= \sum_{j = 1}^q \beta_{(i-1)q + j}(x(t_n), \tilde{y}_m^n) \cdot \varphi_{(i-1)q + j}(h_n).
\end{align*}
\begin{remark}
\textit{The incremental solution operator.} A different method for approximating the local solution operator~$G$ is to construct a DeepONet $G_\theta$ that approximates the following incremental solution operator within the interval $[t_n, t_n+h_n]$:
\begin{align*}
    \tilde{G}_\Delta \left(\tilde{x}(t_n), \tilde{y}_m^n, \lambda \right)(h_n) & \equiv \tilde{x}(t_n + h_n) - x(t_n) \\ &=  \int_{0}^{h_n} f(\tilde{x}(t_n+ s),\tilde{y}^n_m;\lambda) ds,
\end{align*}
and then recover the solution operator $G$ using:
$G \approx G_\theta \left(\tilde{x}(t_n), \tilde{y}_m^n, \lambda \right)(h_n) + x(t_n).$ We have empirically tested both approaches and found that they produce similar experimental results. However, using the incremental operator reduces the training cost.
\end{remark}

\textit{Training the DeepONet.} To train the vector of parameters~$\theta \in \mathbb{R}^p$, we minimize the loss function:
$$\mathcal{L}(\theta) = \frac{1}{N_\text{train}} \sum_{k=1}^{N_\text{train}} \left \|\tilde{G}_{*,k} - G_\theta(x_k(t_n), \tilde{y}_{m,k}^n, \lambda_k)(h_{n,k}) \right \|_2^2,$$
using the dataset of $N_\text{train}:=|\mathcal{D}_\text{train}|$ triplets: $\mathcal{D}_\text{train}:= \{(x_k(t_n), \tilde{y}_{m,k}^n),h_{n,k}, \tilde{G}_{*,k} \}_{k=1}^{N_\text{train}},$ where $\tilde{G}_{*} = \tilde{G} \equiv x(t_n + h_n)$ for the solution operator and $\tilde{G}_{*} = \tilde{G}_\Delta \equiv x(t_n + h_n) - x(t_n) $ for the incremental solution operator.
\vspace{-0.5em}
\subsection{Predicting the Dynamic Response over the Interval~$[0, T]$} \label{sub-sec:DeepONet-scheme}
To predict the dynamic response of the selected SG, we propose the \textit{recursive} DeepONet-based numerical scheme described in Algorithm~\ref{alg:DeepONet-scheme}.

Algorithm~\ref{alg:DeepONet-scheme} requires the following inputs: (i) the trained DeepONet~$G_{\theta^*}$, (ii) the given SG parameters~$\lambda \in \Lambda$, (iii) the initial condition~$x_0 \in \mathcal{X}$, and (iv) the time partition~$\mathcal{P} \subset [0, T]$ of size~$M$ defined as follows:
$\mathcal{P}:~0=t_0 < \ldots < t_M= T,$
where the $n$-th step-size, $h_n:= t_{n+1} - t_n \in (0,h]$, is arbitrary. During the $n$-th recursive step of the algorithm, DeepONet observes (i) the current state~$\tilde{x}(t_n) \in \mathcal{X}$, (ii) the step-size~$h_n$, and (iii) the local discretized input of interacting variables~$\tilde{y}_m^n$ to update the state vector, \ie to compute~$\tilde{x}(t_{n+1})$. The simulated trajectory $\{\tilde{x}(t_n) \in \mathcal{X} : t_n \in \mathcal{P}\}$ is the final output of Algorithm~\ref{alg:DeepONet-scheme}.

\begin{algorithm}[t]
\DontPrintSemicolon
\SetAlgoLined
\textbf{Require:} trained DeepONet~$G_{\theta^*}$, parameters~$\lambda \in \Lambda$, initial state~$x_0 \in \mathcal{X}$, and time partition $\mathcal{P} \subset [0,T]$.\;
Initialize $\tilde{x}(t_0) = x_0$\;
\For{$n = 0,\ldots,M-1$}{
  observe the local interacting variables~$\tilde{y}_m^n$\;
  update the independent variable~$t_{n+1} = t_n + h_n$\;
  update state with the DeepONet's forward pass
  $$\tilde{x}(t_{n+1}) = G_{\theta^*}(\tilde{x}(t_n), \tilde{y}^n_m, \lambda)(h_n).$$
  \vspace{-1.5em}\;}
  \textbf{Return:} simulated trajectory~$\{\tilde{x}(t_n) \in \mathcal{X} : t_n \in \mathcal{P}\}$.\;
 \caption{DeepONet-based Numerical Scheme}
 \label{alg:DeepONet-scheme}
\end{algorithm}

We conclude this section with two remarks. First, an estimation of the cumulative error associated with Algorithm~\ref{alg:DeepONet-scheme} is left for future work. Second, the DeepONet proposed in this section relies entirely on data, which allows it to take advantage of the increasing amount of data available from power grids. However, the power systems community has spent decades developing sophisticated mathematical models to represent dynamic components in power grids. Thus, in the next section, we will design a DeepONet framework that can incorporate both data and mathematical models to predict the dynamic response of SGs.

%% file: 4-residual-learning.tex
In recent years, the amount of data collected by utilities has increased significantly. Meanwhile, the power systems community has developed and optimized sophisticated mathematical models for planning, operating, and controlling the power grid over several decades. Thus, we believe that our proposed DeepONet framework must be able to (i) learn from high-fidelity datasets and (ii) leverage previously developed mathematical models.

In this section, we propose a residual DeepONet, \ie a deep operator network that approximates the residual (or error correction) operator. This residual operator describes the residual dynamics between the SG's true solution operator $\tilde{G}$ and the solution operator resulting from the SG's approximate mathematical model.

Formally, we consider the following non-autonomous IVP based on a previously derived mathematical model of an SG:
\begin{gather}
\begin{aligned} \label{eq:nonautonomous-known}
    \frac{d}{dt}\tilde{x}(t) &= f_\text{approx}(\tilde{x}(t),\tilde{y};\lambda), \qquad t \in [0,T], \\
    \tilde{x}(0) &= x_0,
\end{aligned}    
\end{gather}
where $f_\text{approx}: \mathcal{X} \times \mathcal{Y} \to \mathcal{X}$ is a known vector field that approximates the true vector field~$f$, \ie $f_\text{approx} \approx f$ for a SG. The corresponding \textit{local} solution operator of~\eqref{eq:nonautonomous-known} for any $h_n \in (0,h]$ is:
\begin{align*}
    \tilde{G}_\text{approx} =  \tilde{x}(t_n) + \int_{0}^{h_n} f_\text{approx}(\tilde{x}(t_n+ s),\tilde{y}^n_m;\lambda) ds.
\end{align*}
In practice, we may only have access to this approximate representation of the solution operator, denoted as~$\hat{G}_\text{approx}$. Examples of $\hat{G}_\text{approx}$ include (i) a step of an integration scheme (\eg Runge-Kutta~\cite{iserles2009first}) with variable step-size~$h_n$, (ii) a physics-informed DeepONet~\cite{wang2021long} trained to satisfy~\eqref{eq:nonautonomous-known} locally, or (iii) a DeepONet (see Section~\ref{sec:DeepONet}) trained using a dataset~$\mathcal{D}$ generated by simulating~\eqref{eq:nonautonomous-known}.

Inspired by multi-fidelity schemes~\cite{kim2020multi,wang2020mfpc}, we propose decomposing the local solution operator~$\tilde{G}$ as follows:
\begin{align*}
    \tilde{G} \left(\tilde{x}(t_n), \tilde{y}_m^n, \lambda \right)(h_n) &\equiv \tilde{x}(t_n + h_n) \\
    &= \tilde{x}(t_n) + \int_0^{h_n}f(\tilde{x}(t_n + s), \tilde{y}^n_m,\lambda)ds \\
    &= \tilde{G}_\text{approx}  + \text{``residual dynamics''}.
\end{align*}
Thus, we define the \textit{residual operator}~$G_{\epsilon}$, for any $h_n$, as:
$$G_{\epsilon}\left(\tilde{x}(t_n), \tilde{y}_m^n, \lambda \right)(h_n):=[\tilde{G} - \tilde{G}_\text{approx}]\left(\tilde{x}(t_n), \tilde{y}_m^n, \lambda \right)(h_n).$$
In the above, we have adopted an affine decomposition of the true solution operator $\tilde{G}$. This decomposition simplifies the mathematical analysis of the cumulative error of the proposed residual DeepONet numerical scheme (see Section~\ref{sub-sec:error-bound}). We remark, however, that other decompositions are also possible. For example, we could use the nonlinear multi-fidelity framework~\cite{wang2020mfpc}: $\tilde{G} = G_{\epsilon,\cdot} \circ \tilde{G}_\text{approx} + G_{\epsilon,+}$. We plan to study this framework in our future work.
\vspace{-0.5em}
\subsection{The Residual DeepONet Design} \label{sub-sec:residual-DeepONet}
To approximate the residual operator $G_\epsilon$, we design a residual DeepONet $G_\theta^{\epsilon}$ with trainable parameters $\theta \in \mathbb{R}^p$. $G_\theta^{\epsilon}$ has the same architecture as the data-driven DeepONet (see Fig.~\ref{fig:DeepONet}). Specifically, $G_\theta^{\epsilon}$ consists of a Branch and a Trunk Net, both with the same inputs. The output of the residual DeepONet is defined as follows:
\begin{align*}
    G^{\epsilon}\left(\tilde{x}(t_n), \tilde{y}_m^n, \lambda \right)(h_n) &:= \tilde{e}(t_n + h_n) \in \mathbb{R}^{n_x} \\
    &= x(t_n + h_n) - \tilde{x}_\text{approx}(t_n + h_n) \\
    &= [\tilde{G} - \tilde{G}_\text{approx}]\left(\tilde{x}(t_n), \tilde{y}_m^n, \lambda \right)(h_n),
\end{align*}
We approximate the output of the residual DeepONet using the same dot product as in the data-driven DeepONet case. Finally, we optimize the parameters of the residual DeepONet $G^{\epsilon}_\theta$ by minimizing the loss function:
$$\mathcal{L}(\theta) = \frac{1}{N_\text{train}} \sum_{k=1}^{N_\text{train}} \left \|\tilde{e}_k - G^{\epsilon}_\theta(x_k(t_n), \tilde{y}_{m,k}^n, \lambda_k)(h_{n,k}) \right \|_2^2,$$
using the dataset of $N_\text{train}:=|\mathcal{D}_\text{train}|$ triplets: $\mathcal{D}_\text{train}:= \{(x_k(t_n), \tilde{y}_{m,k}^n),h_{n,k}, \tilde{e}_k(t_n + h_{n,k}) \}_{k=1}^{N_\text{train}}.$
\vspace{-0.5em}
\subsection{Predicting the Dynamic Response over the Interval~$[0,T]$} \label{sub-sec:residual-DeepONet-scheme}
To predict the SG's dynamic response over the interval $[0, T]$, we propose using the recursive residual DeepONet scheme described in Algorithm~\ref{alg:residual-DeepONet}.
\begin{algorithm}[t]
\DontPrintSemicolon
\SetAlgoLined
\textbf{Require:} trained residual DeepONet~$G^{\epsilon}_{\theta^*}$, parameters $\lambda \in \Lambda$, initial state~$x_0 \in \mathcal{X}$, and partition $\mathcal{P} \subset [0,T]$.\;
Initialize $\tilde{x}(t_0) = x_0$\;
\For{$n = 0,\ldots,M-1$}{
  observe the local interacting variables~$\tilde{y}_m^n$\;
  update the independent variable~$t_{n+1} = t_n + h_n$\;
  solve~\eqref{eq:nonautonomous-known} to obtain $\tilde{x}_\text{approx} (t_{n+1})= \hat{G}_\text{approx}\left(\tilde{x}(t_n), \tilde{y}_m^n, \lambda\right)(h_n)$\;
  forward pass of the residual DeepONet
  $$\tilde{e}(t_{n+1}) = G^{\epsilon}_{\theta^*} \left(\tilde{x}(t_n), \tilde{y}^n_m, \lambda \right)(h_n)$$
  \vspace{-1.5em}\;
  update the state vector
  $$\tilde{x}(t_{n+1}) = \tilde{x}_\text{approx}(t_{n+1}) + \tilde{e}(t_{n+1})$$
  \vspace{-1.5em}\;}
  \textbf{Return:} simulated trajectory~$\{\tilde{x}(t_n) \in \mathcal{X} : t_n \in \mathcal{P}\}$.\;
 \caption{Residual DeepONet Numerical Scheme}
 \label{alg:residual-DeepONet}
\end{algorithm}

Algorithm~\ref{alg:residual-DeepONet} takes as input (i) the trained residual DeepONet~$G^{\epsilon}_{\theta^*}$, (ii) the SG's parameters~$\lambda \in \Lambda$, (iii) the initial condition~$x_0 \in \mathcal{X}$, and (iv) the time partition~$\mathcal{P} \subset [0, T]$.

During the $n$th recursive step of the algorithm, we (i) observe the current state~$x(t_n)$ and the local discretized input of interacting variables~$\tilde{y}^n_m$, (ii) solve~\eqref{eq:nonautonomous-known} to approximate the next state vector~$\tilde{x}_\text{approx}(t_n+1)$ (or use a trained physics-informed neural network), (iii) perform a forward pass of residual DeepONet~$G^{\epsilon}_{\theta^*}$ to obtain the predicted error~$\tilde{e}(t_{n+1})$, and (iv) update the state vector via $\tilde{x}(t_{n+1}) = \tilde{x}_\text{approx}(t_{n+1}) + \tilde{e}(t_{n+1})$. Finally, Algorithm~\ref{alg:residual-DeepONet} outputs the simulated trajectory $\{\tilde{x}(t_n) \in \mathcal{X} : t_n \in \mathcal{P}\}$. Let us conclude this section by providing an estimate for the cumulative error of Algorithm~\ref{alg:residual-DeepONet}.
\subsection{Error Bound for the residual DeepONet Numerical Scheme} \label{sub-sec:error-bound}
This section provides an estimate of the cumulative error bound between $x(t_n)$, obtained using the \textit{true} solution operator, and $\hat{x}(t_n)$, obtained using the residual DeepONet numerical scheme detailed in Algorithm~\ref{alg:residual-DeepONet}. To this end, we begin by stating the following assumptions.

\textit{Assumptions.} We assume that the interacting variables, represented by the \textit{input function} $y$, belong to $V \subset C[0,T]$, where $V$ is compact. Additionally, we assume that the vector field $f: \mathcal{X} \times \mathcal{Y} \to \mathcal{X}$ is Lipschitz in~$x$ and $y$, \ie
\begin{align*}
    \|f(x_1,y) - f(x_2, y) \| &\le L ||x_1 - x_2||, \\
    \|f(x,y_1) - f(x, y_2) \| &\le L ||y_1 - y_2||,
\end{align*}
where $L > 0$ is a Lipschitz constant and $x_1$, $x_2$, $y_1$, and $y_2$ are in the appropriate space. Note that engineering systems generally satisfy these assumptions, as $f$ is often differentiable with respect to $x$ and $y$. Nevertheless, we will show empirically (see Section~\ref{sec:numerical-experiments}) that DeepONet can make accurate predictions even when the assumptions above do not hold, such as during a disturbance in the external power grid of the SG.

We now present a lemma that estimates the error bound between $x(t_n)$ obtained from the true solution operator and $\tilde{x}(t_n)$ obtained from the approximate model~\eqref{eq:nonautonomous-approximate}. For simplicity, our analysis assumes $y$ is one-dimensional. However, extending it to multiple inputs is straightforward. 
\begin{lemma} \label{lemma:error-bound-I}
    For any $t_n \in \mathcal{P}$ and $h_n \in (0,h]$, we have
    \begin{align} \label{eq:error-bound-I}
        \|x(t_n) - \tilde{x}(t_n) \| \le \frac{1-r^n}{1-r} \mathcal{E},
    \end{align}
where $r:= e^{Lh}$ and $\mathcal{E} := \max_{n} \{L h_n \kappa_n e^{Lh_n}\}$.
\end{lemma}
\begin{proof}
    For any $h_n \in (0,h]$, let $x(t_{n+1}) \equiv G(x(t_n),y,\lambda)(h_n)$ and $\tilde{x}(t_{n+1}) \equiv G(\tilde{x}(t_n),\tilde{y},\lambda)(h_n)$. Then
    \begin{align*}
        x(t_{n+1}) &= x(t_n) + \int_{t_n}^{t_n + h_n} f(G(x(t_n),y, \lambda)(s), y(s)) ds, \\
        \tilde{x}(t_{n+1}) &= \tilde{x}(t_n) + \int_{t_n}^{t_n + h_n} f(G(\tilde{x}(t_n),\tilde{y}, \lambda)(s), \tilde{y}(s)) ds.
    \end{align*}
We have the following bound for $r_\epsilon := \|x(t_{n+1}) - \tilde{x}(t_{n+1})\|$:
\begin{align*}
     r_\epsilon &\le \|x(t_n) - \tilde{x}(t_n)\| \\ &\quad +  \int_{t_n}^{t_n + h_n} \|f(x(s), y(s)) - f(\tilde{x}(s),\tilde{y}, \lambda)(s), \tilde{y}(s)) \| ds \\
    &\le \|x(t_n) - \tilde{x}(t_n)\| \\ &\quad +   L\int_{t_n}^{t_n + h_n} |y(s) - \tilde{y}(s)|ds + L\int_{t_n}^{t_n + h_n} \|x(s) - \tilde{x}(s)\|ds \\
    &\le \|x(t_n) - \tilde{x}(t_n)\| + L h_n \kappa_n + L\int_{t_n}^{t_n + h_n} \|x(s) - \tilde{x}(s)\|ds.
\end{align*}
In the above, $\kappa_n$ is the local approximation of the input~$y$ within the arbitrary interval~$[t_n, t_n + h_n]$: $\max_{s \in [t_n, t_n + h_n]}~|y(s) - \tilde{y}(s)| \le \kappa_n,$
such that $\kappa \searrow 0$ as the number of sensors $m \nearrow + \infty$. We refer the interested reader to~\cite{lu2021learning} for more details about the above input approximation. Then, using Gronwall's inequality we have
\begin{align*}
    \|x(t_{n+1}) - \tilde{x}(t_{n+1})\| \le \|x(t_n) - \tilde{x}(t_n)\|e^{Lh_n} + Lh_n \kappa e^{Lh_n}.
\end{align*}
Taking $\mathcal{E} := \max_{n} \{L h_n \kappa_n e^{Lh_n}\}$ gives
\begin{align*}
    \|x(t_{n+1}) - \tilde{x}(t_{n+1})\| \le \|x(t_n) - \tilde{x}(t_n)\|e^{Lh_n} + \mathcal{E}.
\end{align*}
The bound~\eqref{eq:error-bound-I} follows due to $x(t_0) = \tilde{x}(t_0) = x_0$.
\end{proof}
Before estimating the error bound between $\tilde{x}(t_n)$ and the residual DeepONet-based prediction $\hat{x}(t_n)$, we need to review the universal approximation theorem of neural networks for high-dimensional functions, as introduced in~\cite{cybenko1989approximation}. To this end, we define a vector-valued continuous function $\varphi: \mathbb{R}^{n_x} \times \mathbb{R}^m \times \mathbb{R}^{n_p} \to \mathbb{R}^{n_x}$ for a given $h_n \in (0,h]$ such that:
$$\varphi(z_n, \tilde{y}^{n}_m, \lambda) = [G - G_\text{approx}](z_n, \tilde{y}_{m}^n, \lambda)(h_n),$$
where $z_n \in \mathbb{R}^{n_x}$. Then, by the universal approximation theorem, for $\epsilon > 0$, there exists $W_1 \in \mathbb{R}^{K \times (n_x + m + n_p)}$, $b_1 \in \mathbb{R}^{K}$, $W_2 \in \mathbb{R}^{n_x \times K}$, and $b_2 \in \mathbb{R}^{n_x}$ such that
\begin{align} \label{eq:nn-approx}
    \left\|\varphi(z_n, \tilde{y}^{n}_m, \lambda) - 
    G^{\epsilon}_{\theta^*}(z_n, \tilde{y}^n_m, \lambda)
     \right\| < \epsilon,
\end{align}
where $G^{\epsilon}_{\theta^*}(z_n, \tilde{y}^n_m, \lambda) =: (W_2 \sigma \left(W_1 \cdot \text{col}(z_n, \tilde{y}^n_m, \lambda) +b_1\right) + b_2).$
We now introduce the following Lemma, which provides an alternative form to describe the local solution operator~$G$ of the approximate system~\eqref{eq:nonautonomous-approximate}.
\begin{lemma}
Consider the local solution operator of the approximate model~\eqref{eq:nonautonomous-approximate}, \ie $G(\tilde{x}(t_n), \tilde{y}^n_m, \lambda)(h_n)$. Then, there exists a function $\Phi: \mathbb{R}^{n_x} \times \mathbb{R}^{m} \times \mathbb{R}^{n_p} \times \mathbb{R} \to \mathbb{R}^{n_x}$, which depends on~$f$, such that
$$\tilde{x}(t_{n+1}) = G(\tilde{x}(t_n), \tilde{y}^{n}_m, \lambda)(h_n) = \Phi(\tilde{x}(t_n), \tilde{y}^n_m, \lambda,h_n),$$
for any $t_n \in \mathcal{P}$ and $h_n \in (0,h]$.
\end{lemma}
We are now ready to provide an estimate for the cumulative error between~$\hat{x}(t_n)$ and $\tilde{x}(t_n)$. 
\begin{lemma} \label{lemma:error-bound-II}
    Assume~$\Phi$ is Lipschitz with respect to the first argument and with Lipschitz constant~$L_{\Phi} > 0$. Suppose the residual DeepONet is well trained so that the neural network architecture satisfies~\eqref{eq:nn-approx}. Then, we have
    $$\|\hat{x}(t_n) - \tilde{x}(t_n)\| \le \frac{1 - L_{\Phi}^n}{1 - L_{\Phi}} \epsilon.$$
\end{lemma}
\begin{proof}
Suppose~$\Phi$ is Lipschitz and the residual DeepONet satisfies the universal approximation theorem of neural networks~\eqref{eq:nn-approx}. Then, for any $t_n \in \mathcal{P}$ and $h_n \in (0,h]$, we have the following bound for $\hat{r}_\epsilon:=\|\hat{x}(t_{n+1}) - \tilde{x}(t_{n+1})\|$:
\begin{align*}
 \hat{r}_\epsilon &= \|G^{\epsilon}_{\theta^*}(\hat{x}(t_n), \tilde{y}^n_m, \lambda) + G_{\text{approx}}(\hat{x}(t_n), \tilde{y}^n_m, \lambda)(h_n) \\ &\qquad - G(\tilde{x}(t_n),\tilde{y}_m^n,\lambda)(h_n) \| \\
 &\le \|G^{\epsilon}_{\theta^*}(\hat{x}(t_n), \tilde{y}^n_m, \lambda) - \varphi(\hat{x}(t_n), \tilde{y}^{n}_m, \lambda) \| \\ & \qquad + \|\Phi(\hat{x}(t_n),\tilde{y}_m^n,\lambda,h_n) - \Phi(\tilde{x}(t_n),\tilde{y}_m^n,\lambda,h_n)\| \\
 &\le \epsilon + L_{\Phi} \|\hat{x}(t_n) - \tilde{x}(t_n)\|.
\end{align*}
The Lemma follows then from $\hat{x}(t_0) = \tilde{x}(t_0) = x_0$.
\end{proof}
The following theorem provides the final estimate for the cumulative error between~$x(t_n)$ obtained using the \textit{true} solution operator~$G$ and $\hat{x}(t_n)$ obtained using the proposed residual DeepONet numerical scheme (see Algorithm~\ref{alg:residual-DeepONet}).
\begin{theorem}
For any~$t_n \in \mathcal{P}$ and $h_n \in (0,h]$, we have
$$\|x(t_n) - \hat{x}(t_n)\| \le \frac{1-r^n}{1-r} \mathcal{E} + \frac{1 - L_{\Phi}^n}{1 - L_{\Phi}} \epsilon.$$
\end{theorem}
The above theorem indicates that error accumulates because of two factors: (i) input approximation and (ii) neural network approximation error. Thus, even if the proposed residual DeepONet generalizes effectively, the final approximation may still be inaccurate. However, we will empirically demonstrate (see Section~\ref{sec:numerical-experiments}) that the proposed DeepONet framework effectively approximates the dynamic response of SGs, even in the extreme case where we locally approximate $y$ using only one sensor ($m=1$), for reasonable values of $h$.

%% file: 5-DAgger.tex
This section considers the problem of a DeepONet-based SG that interacts with a simulator. One significant concern in this problem is the error propagation and accumulation of Algorithms~\ref{alg:DeepONet-scheme} and \ref{alg:residual-DeepONet}. Moreover, this error accumulation increases drastically when working with limited training data. 

This situation of working with limited data may arise when we (i) only have partial knowledge of the state-input space, \ie we only know $\hat{\mathcal{X}} \times \hat{\mathcal{Y}} \subset \mathcal{X} \times \mathcal{Y}$ or (ii) use an expensive and large-scale power grid simulator to solve for the interacting variables $y(t_n)$. Note that the scenario of working with limited data is likely to occur when including control inputs and generator parameters. In particular, sampling from the product space $\mathcal{X} \times \mathcal{Y} \times \mathcal{U} \times \Lambda$ may be prohibitively expensive.

Training a DeepONet-based SG (data-driven or residual) with limited training data may compromise generalization. Moreover, using Algorithms \ref{alg:DeepONet-scheme} or \ref{alg:residual-DeepONet} with DeepONets that fail to generalize effectively may cause error propagation and accumulation. This accumulation of errors can result in distributional drift. That is, the DeepONets will encounter inputs $\{x(t_n),y(t_n)\}$ that were not included in the original training dataset. Observing such inputs can lead to further error propagation and accumulation, putting the long-term prediction accuracy of Algorithms \ref{alg:DeepONet-scheme} and \ref{alg:residual-DeepONet} at risk.

To address distributional drift, we draw inspiration from the behavioral cloning literature~\cite{ross2011reduction} and propose the data aggregation (DAgger) strategy, which is detailed in Algorithm \ref{alg:DAgger}. DAgger is an iterative strategy that trains and fine-tunes DeepONets. In the first iteration of DAgger, we train the data-driven (or residual) DeepONet~$G_\theta$ (or $G^{\epsilon}_\theta$) using the training dataset~$\mathcal{D}_\text{train}$. Then, we use the trained DeepONet~$G_{\theta^*}$ (or $G^{\epsilon}_{\theta^*}$) to rollout Algorithm~\ref{alg:DeepONet-scheme} (or Algorithm~\ref{alg:residual-DeepONet}) $n_\text{rollout}$ times to collect a new dataset of inputs $\mathcal{D} = \{x_k(t_n),y_k(t_n)\}_{k \ge 1}$ generated from the interaction with the simulator. We label the new dataset $\mathcal{D}$ using the true solution operator~$\tilde{G}$, and then aggregate it with the existing training dataset $\mathcal{D}_\text{train}$. We repeat this procedure for $n_\text{iter}$ iterations.
\begin{algorithm}[t]
\DontPrintSemicolon
\SetAlgoLined
\textbf{Require:} Number of rollouts~$n_\text{rollout}$, initial training dataset $\mathcal{D}_\text{train} = \{x_k(t_n),y_k(t_n),h_k, G_{*,k}\}_{k=1}^{N_\text{train}}$, true solution operator~$\tilde{G}$, and partition $\mathcal{P} \subset [0,T_\text{DAgger}]$ (s).\;
\For{$n = 1,\ldots,n_\text{iter}$}{
  train the DeepONet~$G_\theta$ using the dataset~$\mathcal{D}_\text{train}$\;
  run Algorithm~\ref{alg:DeepONet-scheme} or \ref{alg:residual-DeepONet} for $n_\text{rollout}$ using~$G_{\theta^*}$ and~$\mathcal{P}$\;
  collect the dataset of inputs $\mathcal{D} = \{x_k(t_n),y_k(t_n)\}_{k \ge 1}$ visited by~$G_{\theta^*}$\;
  use~$G$ to label the dataset of inputs~$\mathcal{D}$\;
  aggregate the datasets: $\mathcal{D}_\text{train} = \mathcal{D}_\text{train} \cup \mathcal{D}$ \;}
  \textbf{Return:} trained DeepONet~$G_{\theta^*}$.\;
 \caption{Data Aggregation~(DAgger) Algorithm}
 \label{alg:DAgger}
\end{algorithm}

Note that the main idea behind the DAgger Algorithm~\ref{alg:DAgger} is to create a set of inputs that the DeepONet is likely to encounter when interacting with the simulator. Collecting such a set of inputs reduces distributional drift and error accumulation.

%% file: 6-numerical-experiments.tex
\begin{figure}[t!]
\centering
\begin{tikzpicture}[x=0.7pt,y=0.7pt,yscale=-.9,xscale=.9]
\draw  [color={rgb, 255:red, 0; green, 0; blue, 0 }  ,draw opacity=1 ][fill={rgb, 255:red, 184; green, 233; blue, 134 }  ,fill opacity=1 ] (21,56) .. controls (21,42.19) and (32.19,31) .. (46,31) .. controls (59.81,31) and (71,42.19) .. (71,56) .. controls (71,69.81) and (59.81,81) .. (46,81) .. controls (32.19,81) and (21,69.81) .. (21,56) -- cycle ;
\draw  [fill={rgb, 255:red, 248; green, 231; blue, 28 }  ,fill opacity=0.5 ] (100.69,13) -- (215.69,13) -- (215.69,98) -- (100.69,98) -- cycle ;
\draw [line width=2.25]    (71,56) -- (100.67,55.67) ;
\draw [line width=2.25]    (216,58) -- (241,57.9) ;
\draw    (241.36,31.17) -- (241.36,88.17) ;
\draw    (241,47) -- (260.36,38.17) ;
\draw    (241,57.9) -- (260.36,49.07) ;
\draw    (241,67.9) -- (260.36,59.07) ;
\draw    (241,77.9) -- (260.36,69.07) ;
\draw (39,48) node [anchor=north west][inner sep=0.75pt]   [align=left] {$\displaystyle G$};
\draw (112,21) node [anchor=north west][inner sep=0.75pt]   [align=left] {Power Grid};
\draw (142,47) node [anchor=north west][inner sep=0.75pt]   [align=left] {as};
\draw (125,67.4) node [anchor=north west][inner sep=0.75pt]  [font=\large]  {$\infty -$};
\draw (167,67.4) node [anchor=north west][inner sep=0.75pt]   [align=left] {bus};
\end{tikzpicture}
\caption{A synchronous (transient) generator model~$G$ that interacts with the rest of the power grid modeled as an infinite ($\infty$) bus.}
\label{fig:gen-infty-bus}
\vspace{-1.5em}
\end{figure}
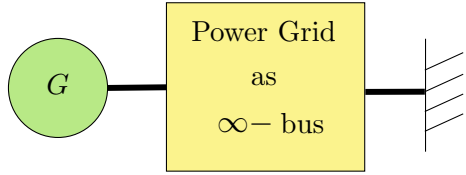

We evaluate the proposed DeepONet frameworks using experiments on the two-axis (transient) generator model~\cite{sauer2017power}, which interacts with a power grid modeled as an infinite bus (see Fig.~\ref{fig:gen-infty-bus}). Our experiments aim to (i) compare the supervised training and generalization efficacy of the DeepONet and residual DeepONet frameworks (Section~\ref{sub-sec:experiment-1}); (ii) analyze the proposed models' sensitivity using different input distributions and dataset sizes (Section~\ref{sub-sec:experiment-2}); and (iii) verify the performance of the DAgger algorithm for different initial conditions and fault sequences~(Section~\ref{sub-sec:experiment-3}).

In all experiments, the proposed DeepONets' predictions are used to solve the network equations of the infinite bus. Additionally, Appendix~\ref{appendix:pst-experiment} provides an experiment that uses DeepONet to shadow the response of a transient SG model interacting with a two-area power grid. We simulate this power grid model using the Power System Toolbox~\cite{chow1992toolbox}. Note that for this experiment in Appendix~\ref{appendix:pst-experiment}, the predictions are not used to solve the power flow and network equations. We begin this section by describing the power grid model used in the experiments, the data generation process, the neural networks, and the training protocol.
\vspace{-0.5em}
\subsection{The Power Grid Model} \label{sub-sec:power-grid-model}
The dynamics describing the two-axis (transient) generator model interacting with an infinite bus are~\cite{sauer2017power}:
\begin{gather} \label{eq:two-axis-model}
    \begin{aligned}
        T'_{d_o} \frac{dE'_q}{dt} &= - E_q' - (X_d - X'_d)I_d + E_{fld} \\
        T'_{q_o} \frac{dE'_d}{dt} &= -E'_d + (X_q - X'_q)I_q\\
        \frac{d\delta}{dt} &= \omega - \omega_s \\
        \frac{2H}{\omega_s} \frac{d\omega}{dt} &= T_M - T_E -D(\omega - \omega_s)
    \end{aligned}
\end{gather}
where $T_E = E'_dI_d + E'_q I_q + (X'_q - X'_d)I_d I_q.$ Here the state vector is $x(t) = (E'_q(t), E'_d(t), \delta(t), \omega(t))^\top$, the parameters are $\{T'_{d_o}, X_d, X'_d, E_{fld}, T'_{q_o}, X_q, X_q', \omega_s, H, T_M, D\}$ (note that the control inputs of the exciter~$E_{fld}$ and the turbine governor~$T_M$ are constant). The input vector is $y(t) = (I_d, I_q)^\top$, which one obtains by solving the network equations of the infinite bus:
\begin{gather} \label{eq:network-equations}
    \begin{aligned}
        0 &= (R_s +R_e)I_d - (X'_q + X_{ep})I_q - E'_d + V_\infty \sin(\delta - \theta_\infty) \\
        0 &= (R_s +R_e)I_q + (X'_d + X_{ep})I_d - E'_q + V_\infty \sin(\delta - \theta_\infty),
    \end{aligned}
\end{gather}
where $\{R_s, R_e, X_{ep}\}$ are parameters and $V_\infty \angle \theta_\infty$ is infinite bus voltage phasor. Note that the network equations~\eqref{eq:network-equations} depend on the predicted state $\delta(t), E'_d(t),$ and $E'_q(t)$. Thus, any DeepONet prediction errors will affect the solution of~\eqref{eq:network-equations}.
\vspace{-0.5em}
\subsection{Data Generation} \label{sub-sec:data-generation}
\textit{Training data.} To generate the training dataset $\mathcal{D}_\text{train}$  for the proposed DeepONet frameworks, we discretize and simulate the dynamics~\eqref{eq:two-axis-model}-\eqref{eq:network-equations}. $\mathcal{D}_\text{train}$ consists of $N_\text{train}$ samples of the form $\{x(t_n),y(t_n),h, G_{*}(x(t_n),y(t_n))(h)\}$. To get the input to the Branch Net, one can perform one of two procedures. The first procedure uniformly samples from the \textit{state-input} distribution. That is, we sample $x(t_n)$ from $\mathcal{X}:=\{x:\delta \in [-\pi, 4\pi], \omega \in [0.0, \pi/2],E'_d \in [-1.0, 1.0],E'_q \in [0.0,1.5]\}$ and $y(t_n)$ from $\mathcal{Y}:=\{y:I_d \in[-1.0, 3.0],I_q \in [-1.0,1.0]\}$. The second procedure uniformly samples $x(t_n)$ from the state space $\mathcal{X}$, and then solves the \textit{network equations} for $y(t_n)$ using the sampled $x(t_n)$.

To get the input for Trunk Net, we sample $h$ uniformly from $[0.0, 0.25]$ (s). The output of DeepONet, $G_{*}$, is obtained as follows. For the incremental operator $G_\Delta$, we first get $x(t_n + h)$ by discretizing and simulating the non-autonomous dynamics~\eqref{eq:two-axis-model} using the sampled values of $x(t_n), y(t_n),$ and $h$. Then, $G_\Delta$ is given by $x(t_n + h) - x(t_n)$. For the residual operator $G_\epsilon$, we obtain the approximate prediction by computing the solution operator of the approximate model, \ie $G_\text{approx}$ using $x(t_n), y(t_n),$ and $h$. The approximate model in this experiment is the same as the dynamics~\eqref{eq:two-axis-model}, except for the angular velocity equation, which reads as follows:
$$ \frac{2H}{\omega_s} \frac{d\omega}{dt} = T_M - T_E - \beta D(\omega - \omega_s).$$
We use the parameter $\beta = 0.5$ to adjust the model's accuracy. Using the approximate and true solution operators, we then compute the residual operator via $G_\epsilon \equiv G - G_\text{approx} = x(t_n + h) - x_\text{approx}(t_n + h)$.

\textit{Test data.} We generate two test datasets of $N_\text{test}=500$ solution trajectories $\{x^{(i)}(t): t \in \mathcal{P} \}_{i=1}^{N\text{test}}$ for a given time partition $\mathcal{P}$. The \textit{first} test dataset is obtained as follows. Let $x^* = (\delta^*, \omega^*, E'^{*}_d, E'^{*}_q)^\top$ denote the equilibrium point of~\eqref{eq:two-axis-model}. Then, one obtains each test trajectory from an initial condition $x_o \in \mathcal{X}_o :=\{x: \delta=\delta^*, \omega = \gamma \omega^*, E'_d=E'^{*}_d, E'_q=E'^{*}_q\},$ where $\gamma$ is uniformly sampled from the interval $[0.2,1.5]$ (s). 

Each trajectory in the \textit{second} test dataset starts from the equilibrium $x_o = x^*$. At time $t_f=1.0$ (s), a fault is simulated by changing the impedance of the infinite bus. The fault then is cleared by restoring the original impedance value at time $t_f + \Delta t_f$. The duration of the fault $\Delta t_f$ varies across trajectories and is uniformly sampled from the interval $[0.05, 1.0]$ (s).
\vspace{-0.7em}
\subsection{Neural Nets and Training Protocol} \label{sub-sec:NNs-training}
\textit{Neural Nets.} We implemented the proposed frameworks using PyTorch. For the Branch and Trunk Nets, we (i) used the modified fully-connected network architecture (also used in our previous work~\cite{moya2022dae,moya2022deeponet}), (ii) employed leaky ReLU activation functions, and (iii) conducted a simple hyperparameter optimization routine to identify the best architectures.

\textit{Training.}We trained the neural networks using supervised learning and the proposed DAgger algorithm (see Section~\ref{sec:DAgger}). The DeepONet parameters were optimized using Adam (with learning rate $\eta = 5 \times 10^{-3}$). We also used a scheduler that reduces the learning rate whenever the training reaches a plateau. Finally, we normalized the inputs and outputs during training and testing.
\vspace{-0.5em}
\subsection{Experiment 1. Supervised Training} \label{sub-sec:experiment-1}
This experiment evaluates the efficiency and effectiveness of data-driven and residual DeepONets in supervised training. To this end, we trained the proposed frameworks on $N_\text{train}= 2000$ samples (80\% training, 20\% validation), with the inputs to the Branch Net obtained by sampling the state-input distribution $\mathcal{X} \times \mathcal{Y}$. We ran supervised training for $N_\text{epochs} = 2000$ epochs, validating after each epoch.

\begin{figure}[t!]
\centering
\begin{subfigure}[b]{0.4\textwidth}
\centering
\includegraphics[width=0.99\textwidth]{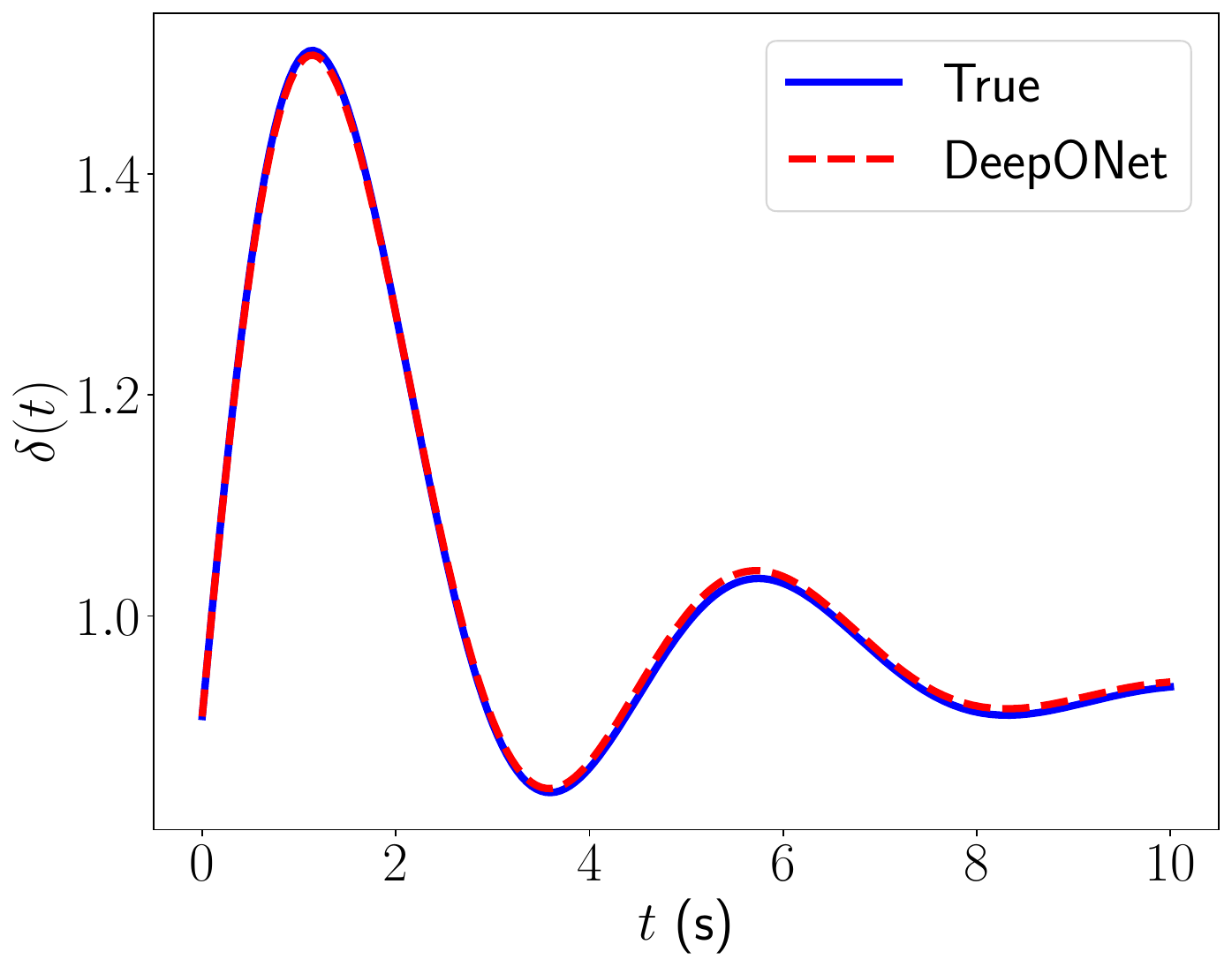}
\end{subfigure}
\begin{subfigure}[b]{0.4\textwidth}
\centering
\includegraphics[width=0.99\textwidth]{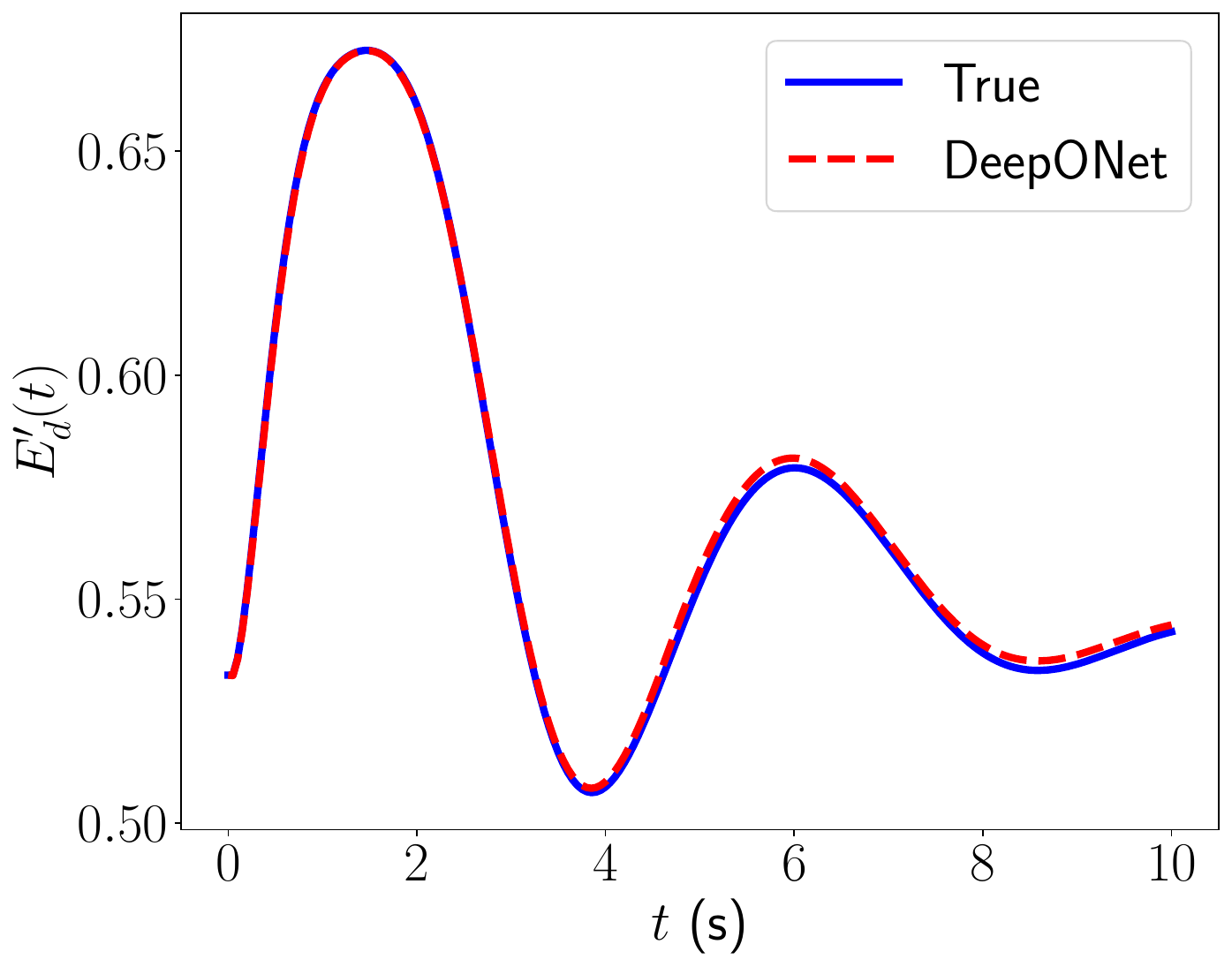}
\end{subfigure}
\caption{Comparison of the data-driven DeepONet prediction with the true trajectory of the SG selected states $(\delta(t), E_d'(t))$ within the uniform partition $\mathcal{P} \subset [0,10]$ (s) of constant step size $h = 0.05$.}
\label{fig:comparison-states-DD-DeepONet}
\vspace{-1em}
\end{figure}

\begin{figure}[t!]
\centering
\begin{subfigure}[b]{0.4\textwidth}
\centering
\includegraphics[width=0.99\textwidth,]{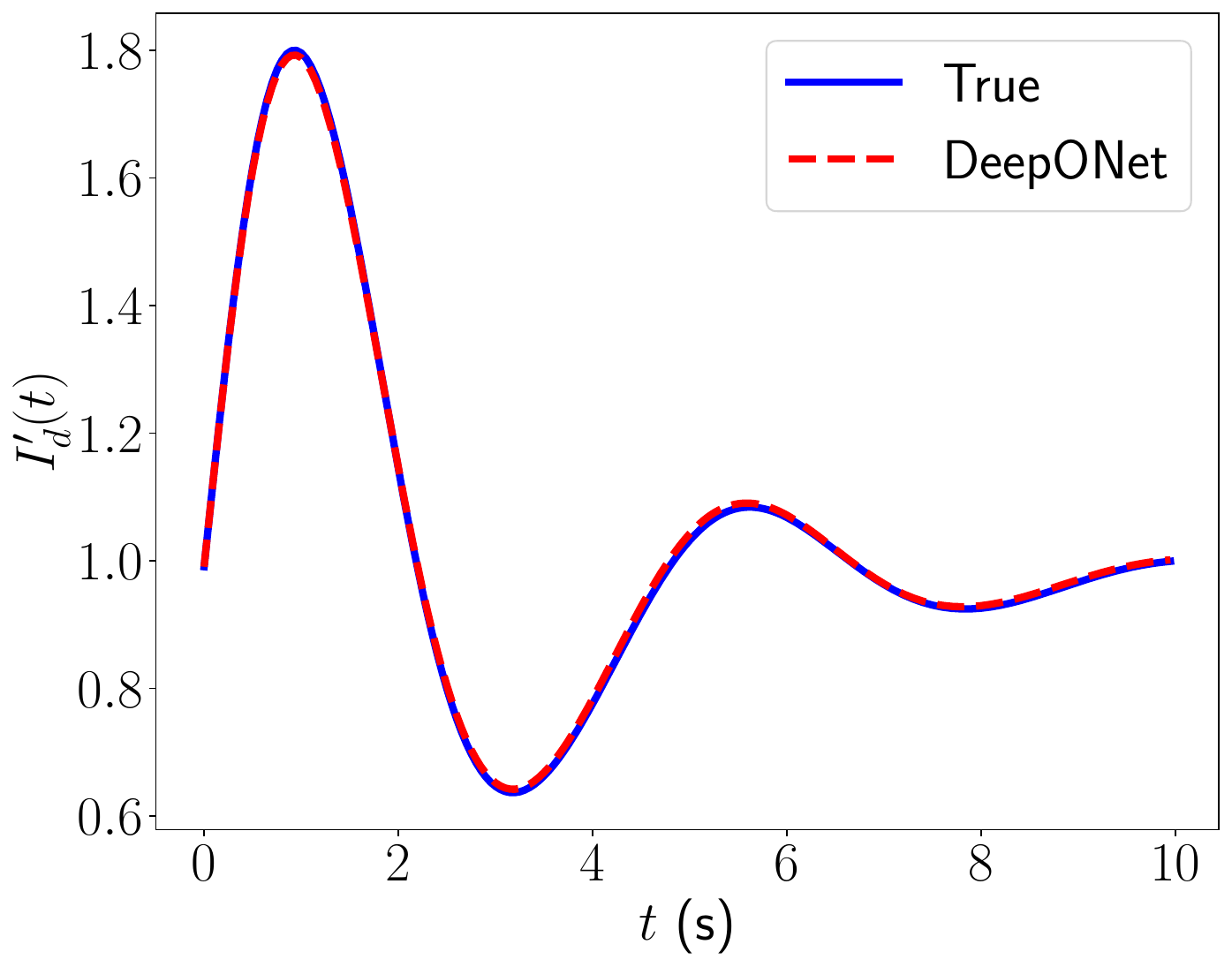}
\end{subfigure}
\begin{subfigure}[b]{0.4\textwidth}
\centering
\includegraphics[width=0.99\textwidth]{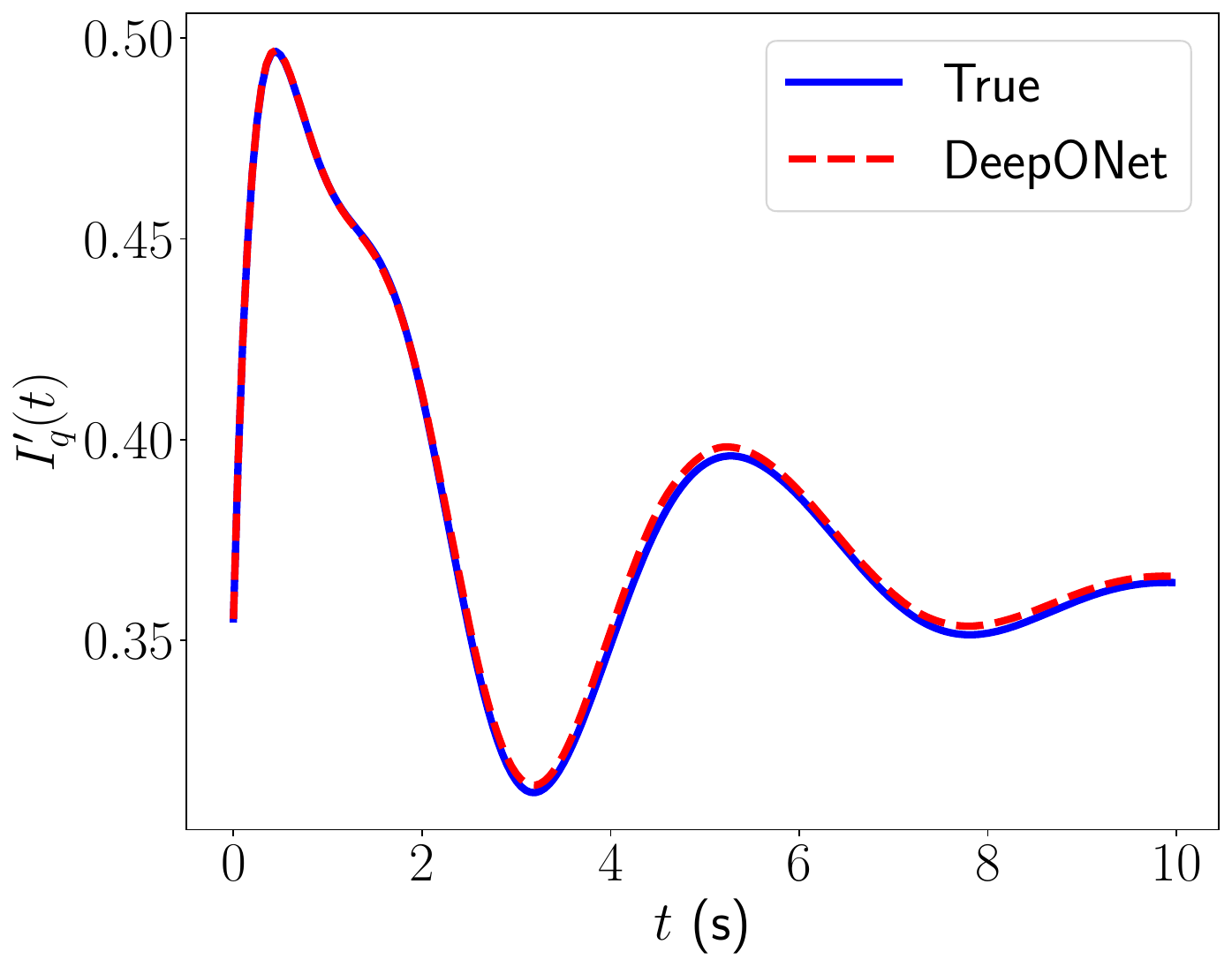}
\end{subfigure}
\caption{Comparison of the resulting network input $y(t) = (I_d'(t), I_q'(t))^\top$ for the data-driven DeepONet trained within the uniform partition $\mathcal{P} \subset [0,10]$ (s) of constant step size $h = 0.05$.}
\label{fig:comparison-inputs-DD-DeepONet}
\vspace{-1em}
\end{figure}

\begin{figure}[t!]
\centering
\begin{subfigure}[b]{0.4\textwidth}
\centering
\includegraphics[width=0.99\textwidth]{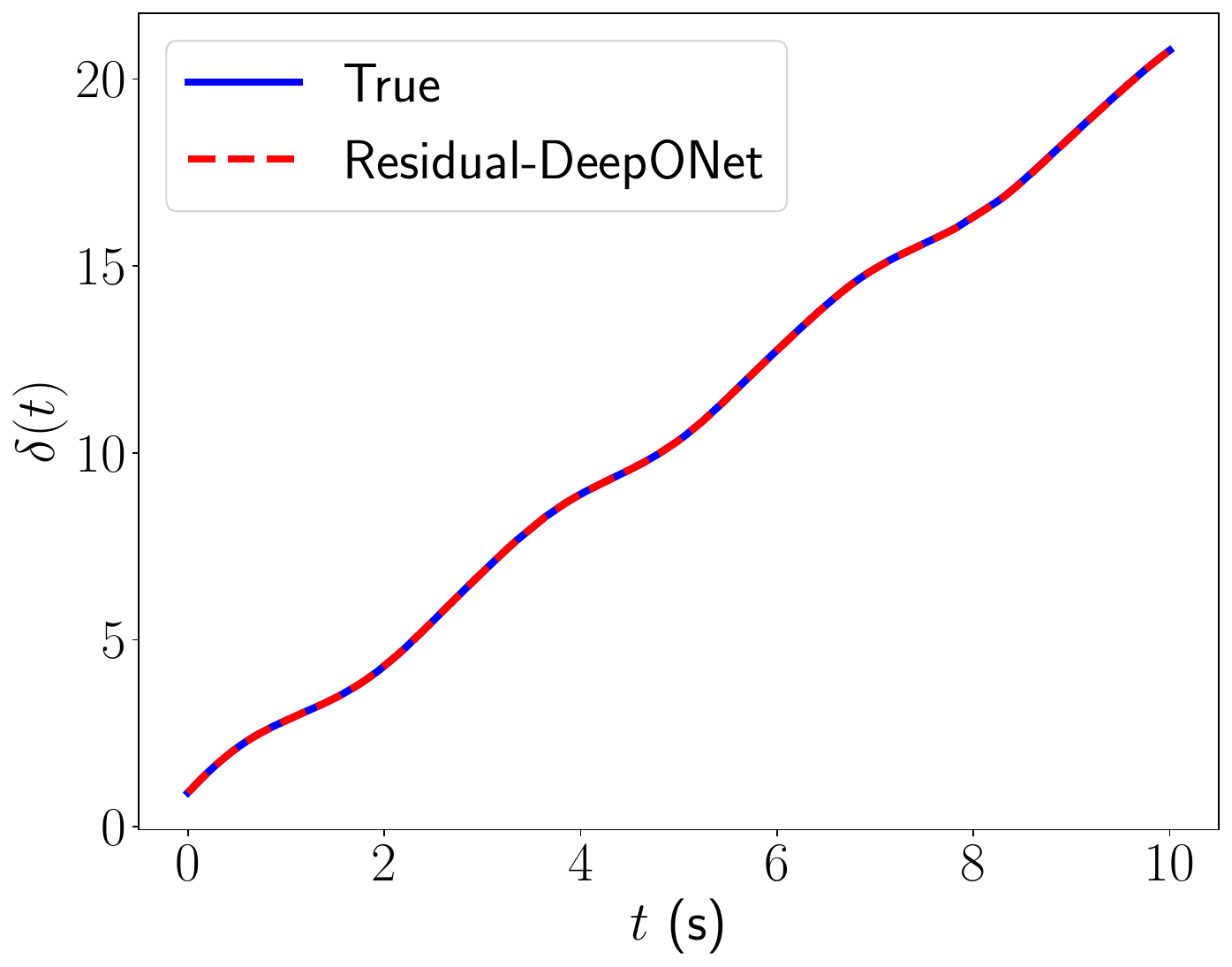}
\end{subfigure}
\begin{subfigure}[b]{0.4\textwidth}
\centering
\includegraphics[width=0.99\textwidth]{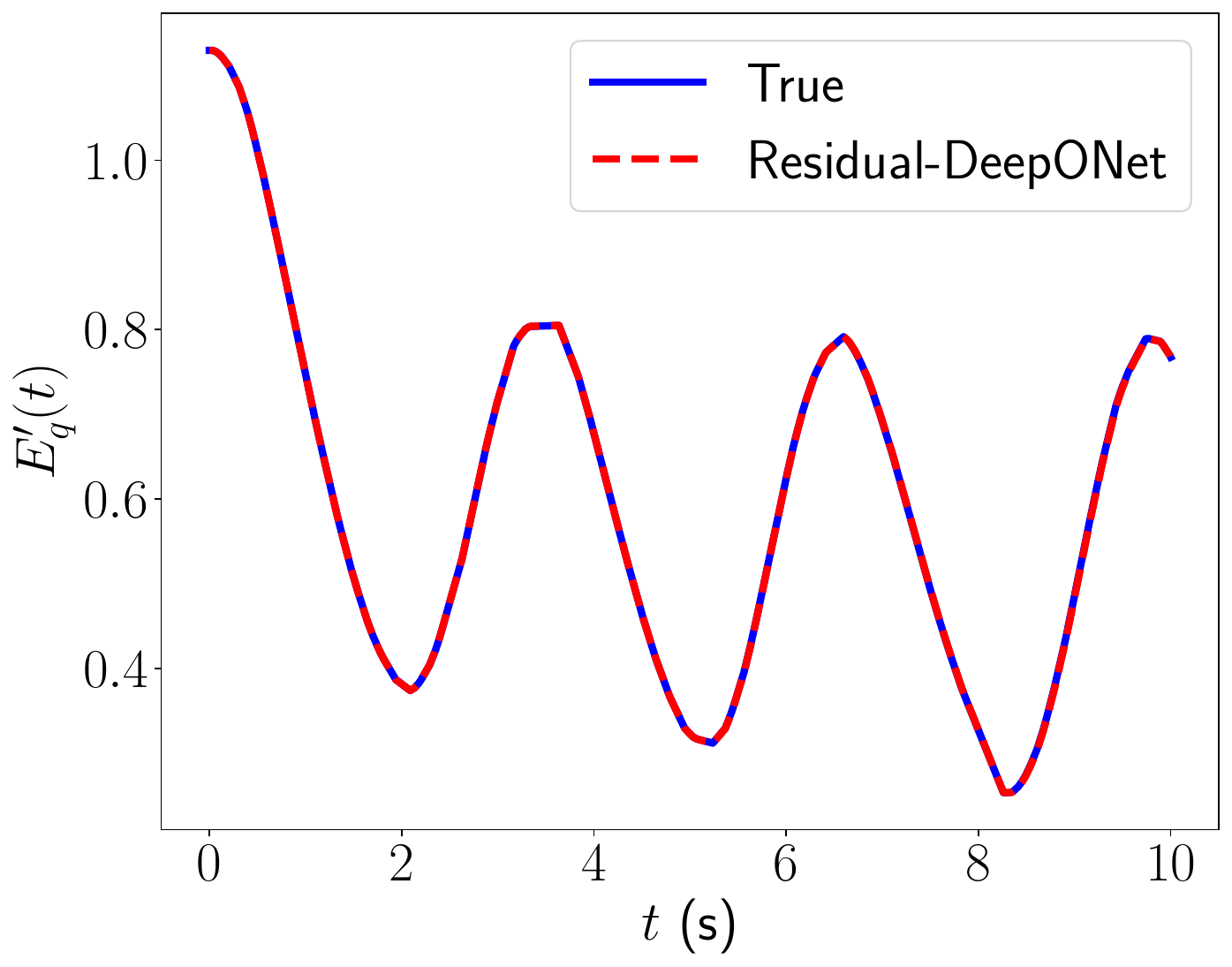}
\end{subfigure}
\caption{Comparison of the residual DeepONet prediction with the true trajectory of the SG selected states $(\delta(t), E_q'(t))$ within the irregular partition $\mathcal{P} \subset [0,10]$ (s) of size $200$ (random points).}
\label{fig:comparison-states-residual-DeepONet}
\vspace{-1em}
\end{figure}

\begin{figure}[t!]
\centering
\begin{subfigure}[b]{0.4\textwidth}
\centering
\includegraphics[width=0.99\textwidth]{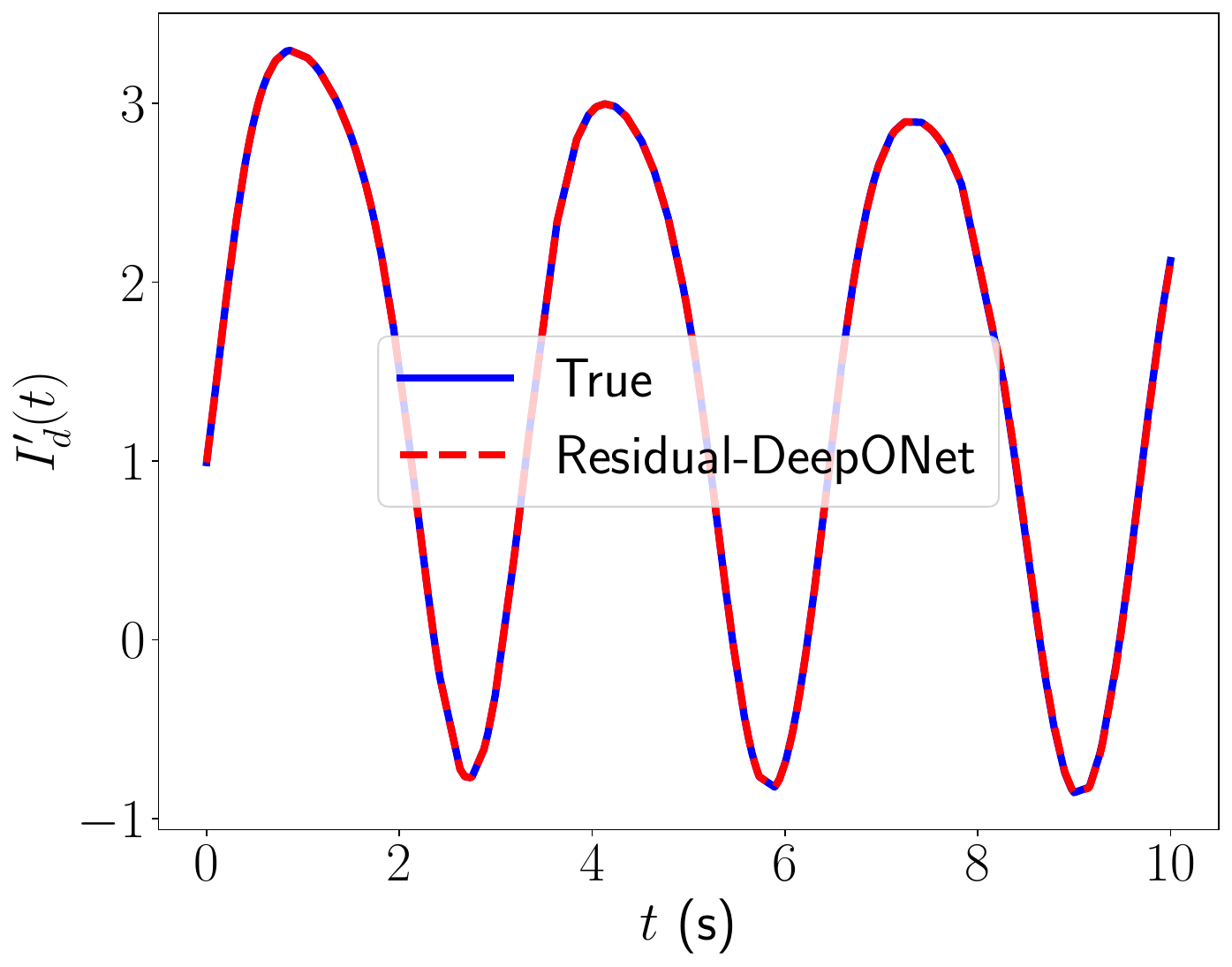}
\end{subfigure}
\begin{subfigure}[b]{0.4\textwidth}
\centering
\includegraphics[width=0.99\textwidth]{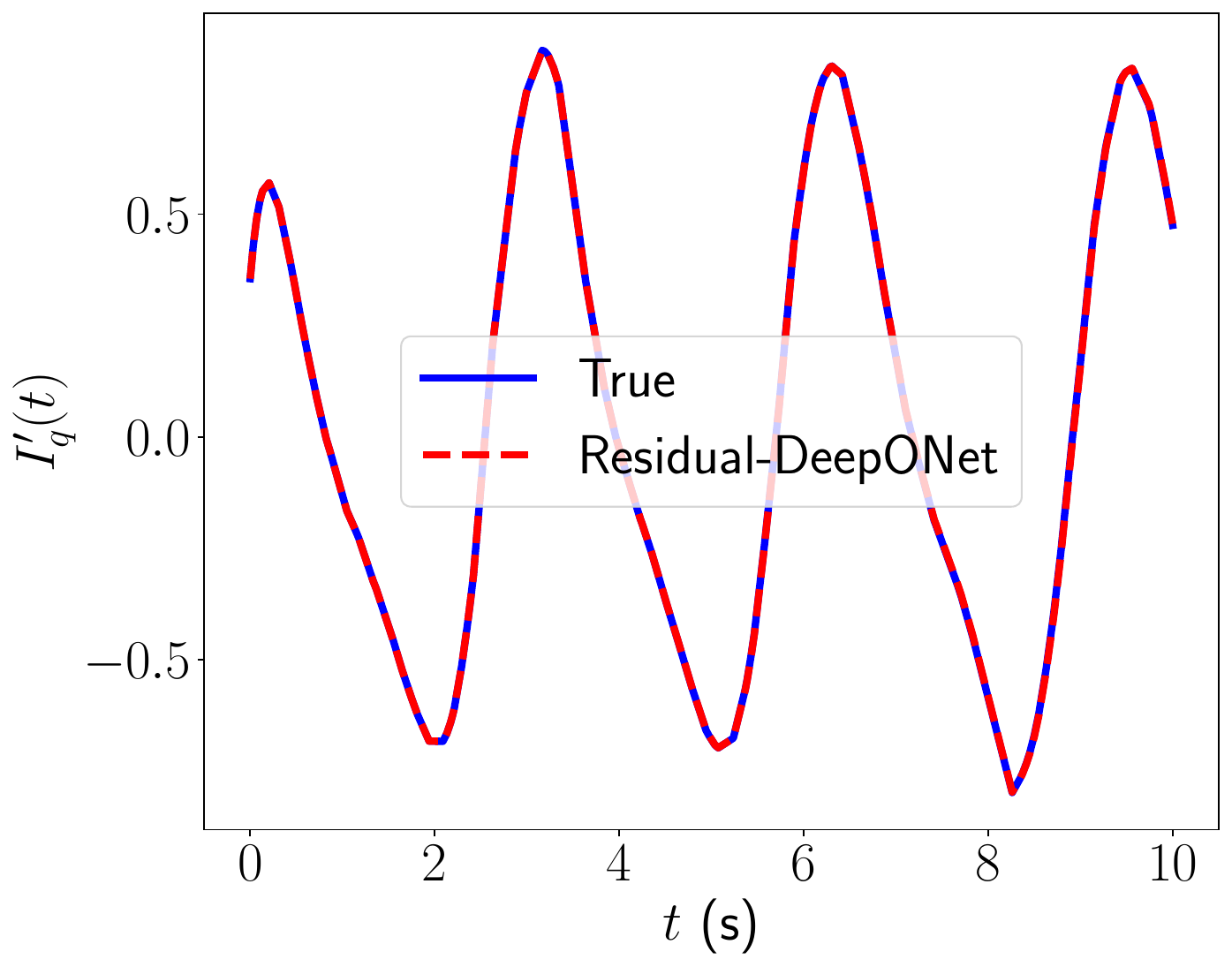}
\end{subfigure}
\caption{Comparison of the resulting network input $y(t) = (I_d'(t), I_q'(t))^\top$ for the residual DeepONet trained within the irregular partition $\mathcal{P} \subset [0,10]$ (s) of size $200$ (random points).}
\label{fig:comparison-inputs-residual-DeepONet}
\vspace{-1em}
\end{figure}

We used the trained DeepONet frameworks to test their generalization outside the training dataset. Random trajectories were selected from the test dataset and the predicted DeepONet results were compared against the ground truth. Fig.~\ref{fig:comparison-states-DD-DeepONet} and Fig.~\ref{fig:comparison-inputs-DD-DeepONet} (resp. Fig.~\ref{fig:comparison-states-residual-DeepONet} and Fig.~\ref{fig:comparison-inputs-residual-DeepONet}) show the prediction for selected state trajectories and network responses for the data-driven (resp. residual) DeepONet over a uniform (resp. irregular) partition $\mathcal{P} \subset [0,10]$ (s). Both frameworks accurately predict state trajectories and do not introduce errors when solving network equations. Furthermore, the residual DeepONet can effectively learn unstable and oscillatory trajectories, which is often a challenging task.

To study the predictive and generalization capabilities of the proposed frameworks, we computed the mean and standard deviation of the $L_2-$relative error for all $500$ test trajectories. Tables~\ref{table:DD-DeepONet} and~\ref{table:residual-DeepONet} show the results for the data-driven and residual DeepONet, respectively. We also compared our results with a trained, fully-connected neural network (FNN), which is often used in model-based reinforcement learning to learn a surrogate of the environment and predict the next state. Table~\ref{table:MLP} shows the $L_2$-relative errors between the FNN prediction and the ground truth for all test trajectories.
\begin{table}[t]
\centering
\begin{tabular}{ l | c  c  c  c c c}
\hline
& $\delta(t)$ & $\omega(t)$ & $E_d'(t)$ & $E_q'(t)$ & $I_d'(t)$ & $I_q'(t)$\\
\hline
mean $L_2$ \% & 1.3 \% & 0.32 \% & 2.9 \% & 1.2 \% & 2.2 \% & 3.5 \% \\
st. dev. $L_2$ \% & 3.5 \% & 0.77 \% & 8.6 \% & 3.6 \% & 5.1 \% & 10.3 \% \\
\hline
\end{tabular}
\caption{The mean and standard deviation (st. dev.) of the $L_2$-relative error between the predicted (data-driven DeepONet) and the actual response of the (i) synchronous generator state and (ii) network's input for $500$ initial conditions sampled from the set~$\mathcal{X}_o$ and within the uniform partition $\mathcal{P} \subset [0,10]$ (s) of constant step size $h = 0.05$.}
\label{table:DD-DeepONet}
\vspace{-0.5em}
\end{table}

\begin{table}[t]
\centering
\begin{tabular}{ l | c  c  c  c c c}
\hline
& $\delta(t)$ & $\omega(t)$ & $E_d'(t)$ & $E_q'(t)$ & $I_d'(t)$ & $I_q'(t)$\\
\hline
mean $L_2$ \% & 0.2 \% & 0.05 \% & 0.6 \% & 0.22 \% & 0.4 \% & 0.7 \% \\
st. dev. $L_2$ \% & 0.5 \% & 0.13 \% & 1.6 \% & 0.6 \% & 1.1 \% & 2.1 \% \\
\hline
\end{tabular}
\caption{The mean and standard deviation (st. dev.) of the $L_2$-relative error between the predicted (residual DeepONet) and the actual response of the (i) synchronous generator state and (ii) network's input for $500$ initial conditions sampled from the set~$\mathcal{X}_o$ and within the irregular partition $\mathcal{P} \subset [0,10]$ (s) of size $200$ (distinct random points).}
\label{table:residual-DeepONet}
\vspace{-0.5em}
\end{table}

\begin{table}[t]
\centering
\begin{tabular}{ l | c  c  c  c c c}
\hline
& $\delta(t)$ & $\omega(t)$ & $E_d'(t)$ & $E_q'(t)$ & $I_d'(t)$ & $I_q'(t)$\\
\hline
mean $L_2$ \% & 22.5\% & 4.0\% & 37.8\% & 15.3\% & 25.8\% & 36.5\% \\
st. dev. $L_2$ \% & 16.3\% & 4.7\% & 46.9\% & 13.6\% & 23.9\% & 45.9\% \\
\hline
\end{tabular}
\caption{The mean and standard deviation (st. dev.) of the $L_2$-relative error between the predicted (feed-forward neural network) and the actual response of the (i) synchronous generator state and (ii) network's input for $500$ initial conditions sampled from the set~$\mathcal{X}_o$ and within a uniform partition $\mathcal{P} \subset [0,10]$ (s) of constant step size $h=0.05$.}
\label{table:MLP}
\vspace{-0.5em}
\end{table}

The results indicate that: (i) FNN cannot predict the SG's dynamic response, and its errors accumulate over time. In practice, the reinforcement learning community uses ensembles and uncertainty quantification~\cite{chua2018deep} to reduce this error accumulation. However, these methods increase the computational cost. (ii) Our proposed frameworks can control the error accumulation without increasing computational costs. For example, data-driven DeepONet keeps the mean $L_2$ error for all states below $5$\%, which can be further reduced by increasing the number of training samples or using the DAgger algorithm. (iii) Finally, the residual DeepONet yielded the best results, with a mean $L_2$ error always below $1$\%. This suggests that incorporating prior information, such as the underlying physics of the problem, improves generalization even with limited training examples.
\subsection{Experiment 2. Sensitivity} \label{sub-sec:experiment-2}
In this experiment, we test the generalization capabilities of the proposed frameworks using the Branch Net's training inputs $(x(t_n),y(t_n))$ obtained by sampling $x(t_n)$ from $\mathcal{X}$ and solving the network equations \eqref{eq:network-equations} for $y(t_n)$. We trained the DeepONets as in Experiment~1 and computed the $L_2$-relative errors for all $500$ test trajectories. The results for the data-driven (resp. residual) DeepONet are shown in Table~\ref{table:DD-DeepONet-network} (resp. Table~\ref{table:residual-DeepONet-network}). These results show that training using inputs obtained by solving \eqref{eq:network-equations} has similar $L_2$ errors as training using inputs sampled from the input space $\mathcal{Y}$. However, we remark that training using the (i) state-input distribution requires knowledge of the spaces $\mathcal{X}$ and $\mathcal{Y}$, and (ii) network equations require solving the entire power grid dynamics, which may be costly for large-scale models.

\begin{table}[t]
\centering
\begin{tabular}{ l | c  c  c  c c c}
\hline
& $\delta(t)$ & $\omega(t)$ & $E_d'(t)$ & $E_q'(t)$ & $I_d'(t)$ & $I_q'(t)$\\
\hline
mean $L_2$ \% & 1.96\% & 0.41\% & 4.94\% & 1.73\% & 3.70\% & 5.30\% \\
st. dev. $L_2$ \% & 2.91\% & 0.66\% & 8.42\% & 2.60\% & 5.25\% & 10.16\% \\
\hline
\end{tabular}
\caption{The mean and standard deviation (st.dev.) of the $L_2$-relative error between the predicted (data-driven DeepONet trained using the network equations~\eqref{eq:network-equations}) and the actual response of the (i) synchronous generator state and (ii) network's input for $500$ initial conditions sampled from the set~$\mathcal{X}_o$ and within the uniform partition $\mathcal{P} \subset [0,10]$ (s) of constant step size $h = 0.05$.}
\label{table:DD-DeepONet-network}
\vspace{-0.5em}
\end{table}

\begin{table}[t]
\centering
\begin{tabular}{ l | c  c  c  c c c}
\hline
& $\delta(t)$ & $\omega(t)$ & $E_d'(t)$ & $E_q'(t)$ & $I_d'(t)$ & $I_q'(t)$\\
\hline
mean $L_2$ \% & 0.28\% & 0.06\% & 0.76\% & 0.30\% & 0.56\% & 0.92\% \\
st. dev. $L_2$ \% & 0.68\% & 0.17\% & 2.04\% & 0.77\% & 1.38\% & 2.66\% \\
\hline
\end{tabular}
\caption{The mean and standard deviation (st. dev.) of the $L_2$-relative error between the predicted (residual DeepONet trained using the network equations~\eqref{eq:network-equations}) and the actual response of the (i) synchronous generator state and (ii) network's input for $500$ initial conditions sampled from the set~$\mathcal{X}_o$ and within the uniform partition $\mathcal{P} \subset [0,10]$ (s) of constant step size $h = 0.05$.}
\label{table:residual-DeepONet-network}
\vspace{-0.5em}
\end{table}

\begin{figure}[t!]
\centering
\begin{subfigure}[b]{0.4\textwidth}
\centering
\includegraphics[width=0.99\textwidth]{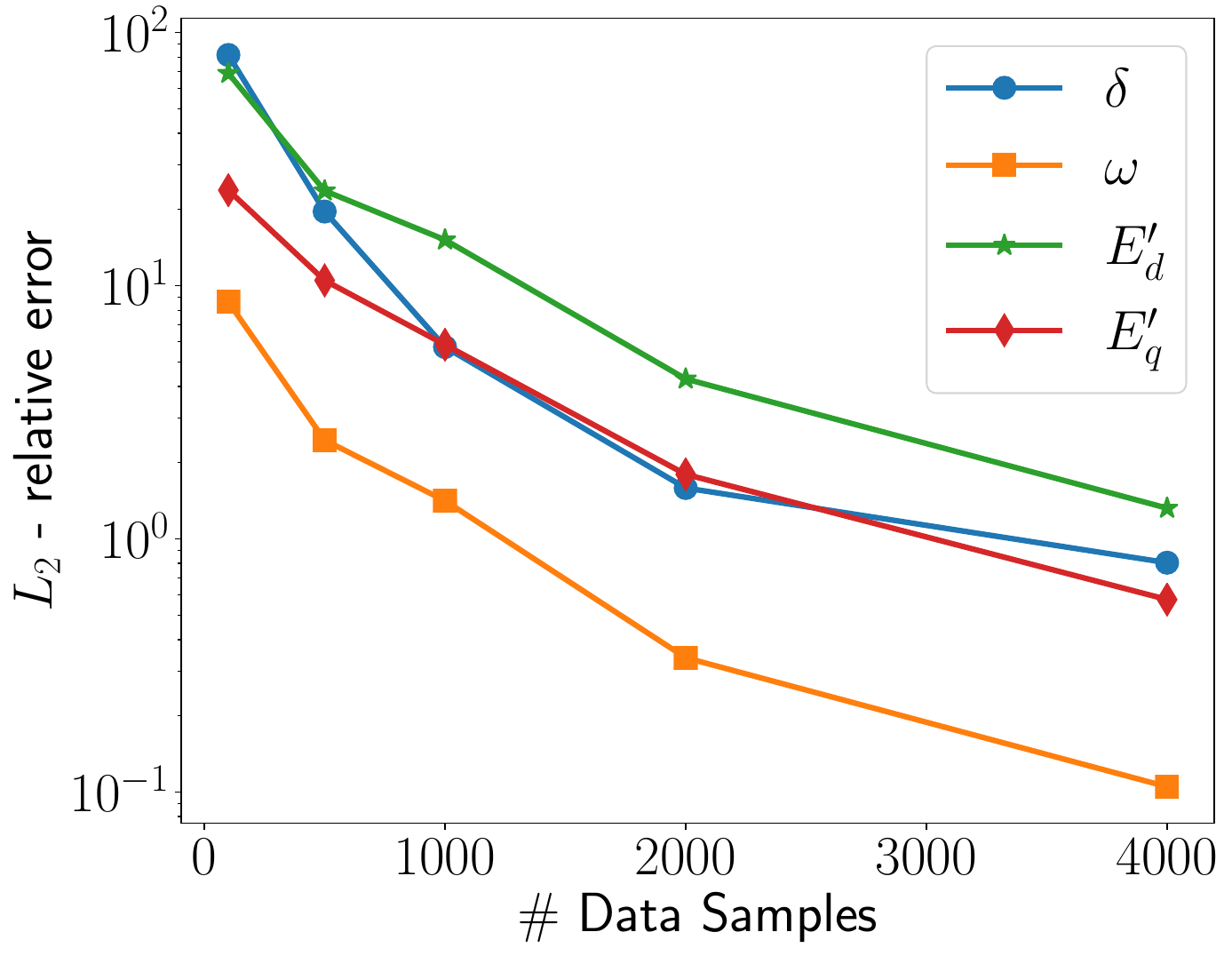}
\end{subfigure}
\begin{subfigure}[b]{0.4\textwidth}
\centering
\includegraphics[width=0.99\textwidth]{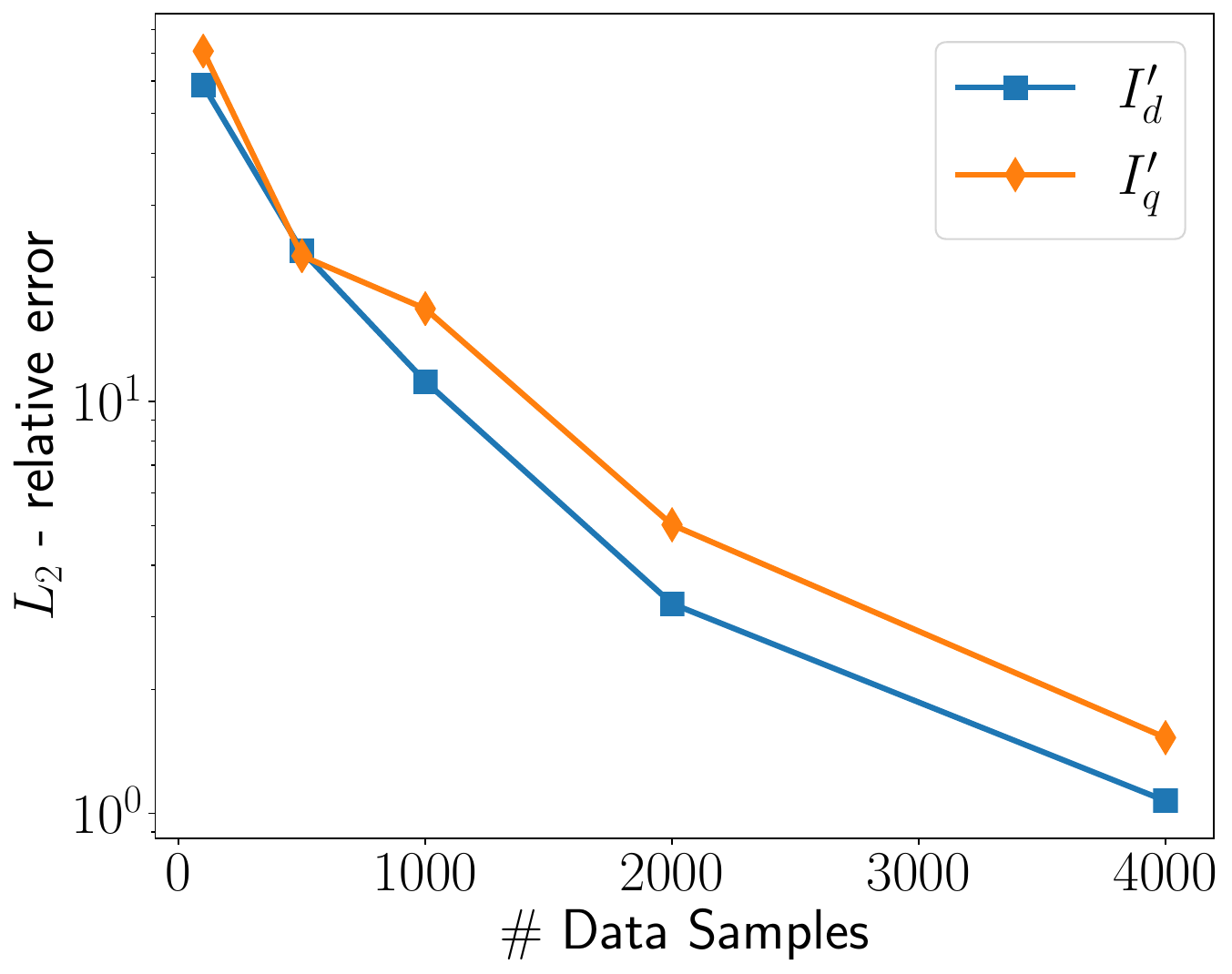}
\end{subfigure}
\caption{The mean $L_2$-relative error~\% of $500$ predicted trajectories with data-driven DeepONet as a function of the number of training samples for the (\textit{left}) generator states $x(t) = (\delta(t), \omega(t), E_d'(t), E_q'(t))^\top$ and (\textit{right}) network inputs $y(t) = (I_d'(t), I_q'(t))^\top$.}
\label{fig:comparison-dd-L2-rel-error-nData}
\vspace{-1em}
\end{figure}

\begin{figure}[t!]
\centering
\begin{subfigure}[b]{0.4\textwidth}
\centering
\includegraphics[width=0.99\textwidth]{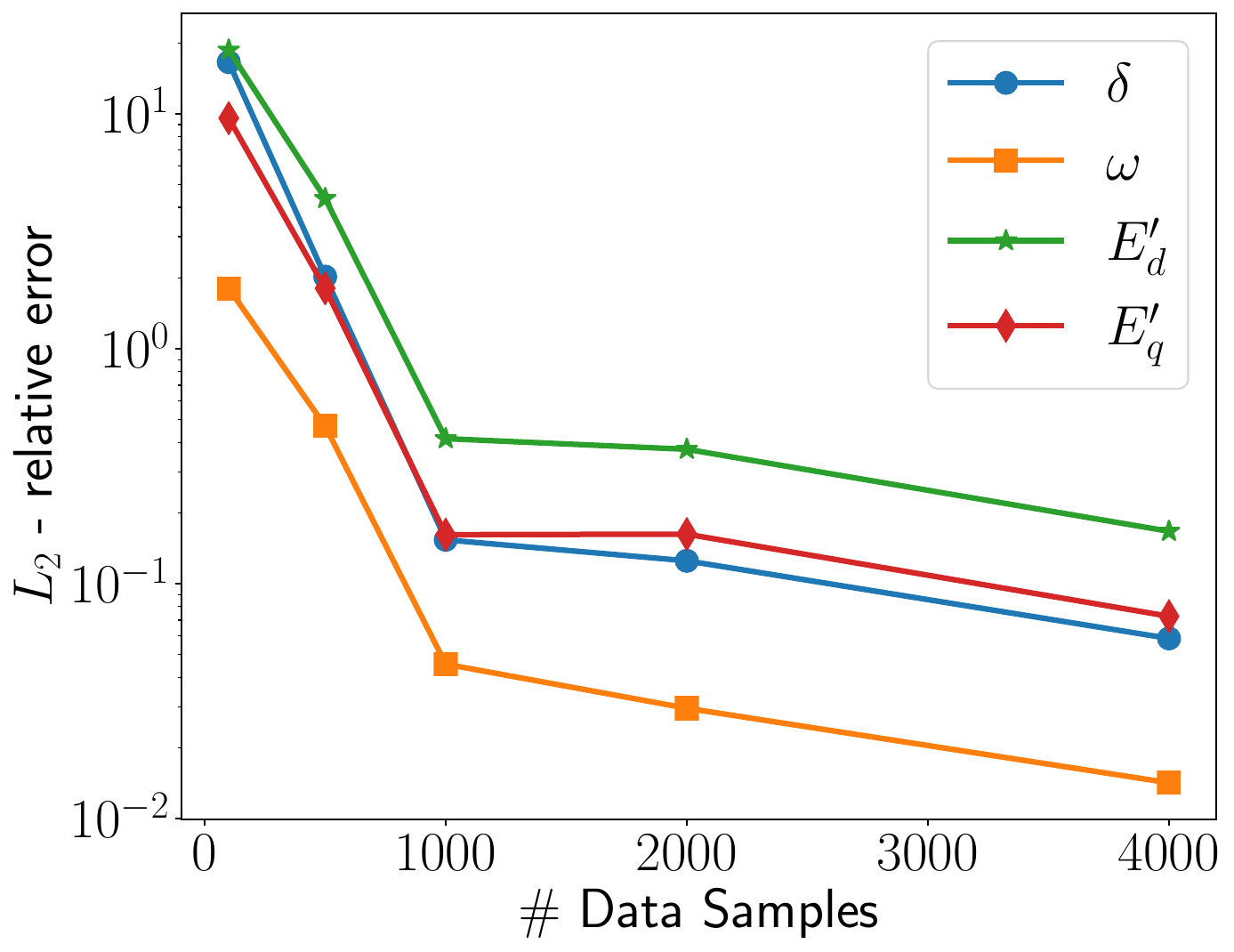}
\end{subfigure}
\begin{subfigure}[b]{0.4\textwidth}
\centering
\includegraphics[width=0.99\textwidth]{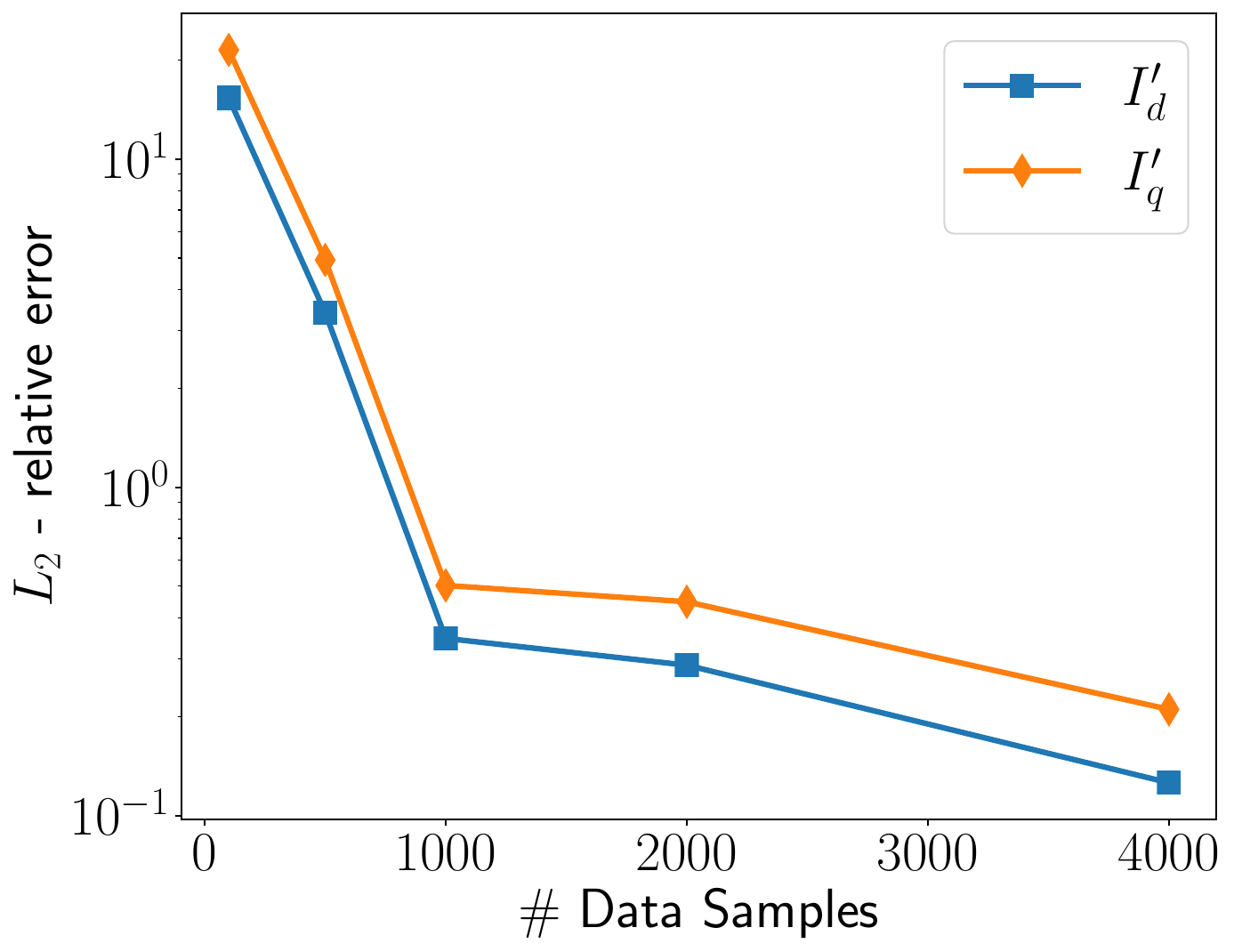}
\end{subfigure}
\caption{The mean $L_2$-relative error~\% of $500$ predicted trajectories with residual DeepONet as a function of the number of training samples for the (\textit{left}) generator states $x(t) = (\delta(t), \omega(t), E_d'(t), E_q'(t))^\top$ and (\textit{right}) network inputs $y(t) = (I_d'(t), I_q'(t))^\top$.}
\label{fig:comparison-residual-L2-rel-error-nData}
\vspace{-1em}
\end{figure}

We now analyze how $L_2$-relative errors for all the 500 test trajectories change with the number of training samples. For simplicity, we let the number of samples vary from the set $\{100, 500, 1000, 2000, 4000\}$. Fig.~\ref{fig:comparison-dd-L2-rel-error-nData} (resp. Fig.~\ref{fig:comparison-residual-L2-rel-error-nData}) shows how the $L_2$-relative errors (for state and network trajectories) change with an increase in training data samples for the data-driven (resp. residual) DeepONet. The results demonstrate that increasing the number of training samples reduces $L_2$-relative errors and error accumulation. We also observe that residual DeepONet (using prior information) outperforms data-driven DeepONet in terms of generalization and error accumulation, regardless of training dataset size.

\subsection{Experiment 3. The DAgger Algorithm} \label{sub-sec:experiment-3}
This experiment uses the DAgger algorithm (see Section~\ref{sec:DAgger}) to train the residual DeepONet framework. DAgger is useful when sampling the state-input space $\mathcal{X} \times \mathcal{Y}$ is too expensive, and supervised learning may accumulate and propagate errors. To simulate this situation, we only collected $N_\text{train} = 100$ data samples for supervised learning and used $n_\text{iters}=5$ iterations of DAgger. Each DAgger iteration predicted $10$ rollouts sampled from different initial conditions $x(0) \in \mathcal{X}_o$ and over a uniform partition $\mathcal{P} \subset [0.0,5.0]$ (s) of step size $h = 0.05$. After training with DAgger, we tested the residual DeepONet's generalization using a test dataset of 500 disturbance trajectories. Note that this test scenario is challenging because DAgger never observed a disturbance trajectory during training. 

\begin{figure}[t!]
\centering
\begin{subfigure}[b]{0.4\textwidth}
\centering
\includegraphics[width=0.99\textwidth]{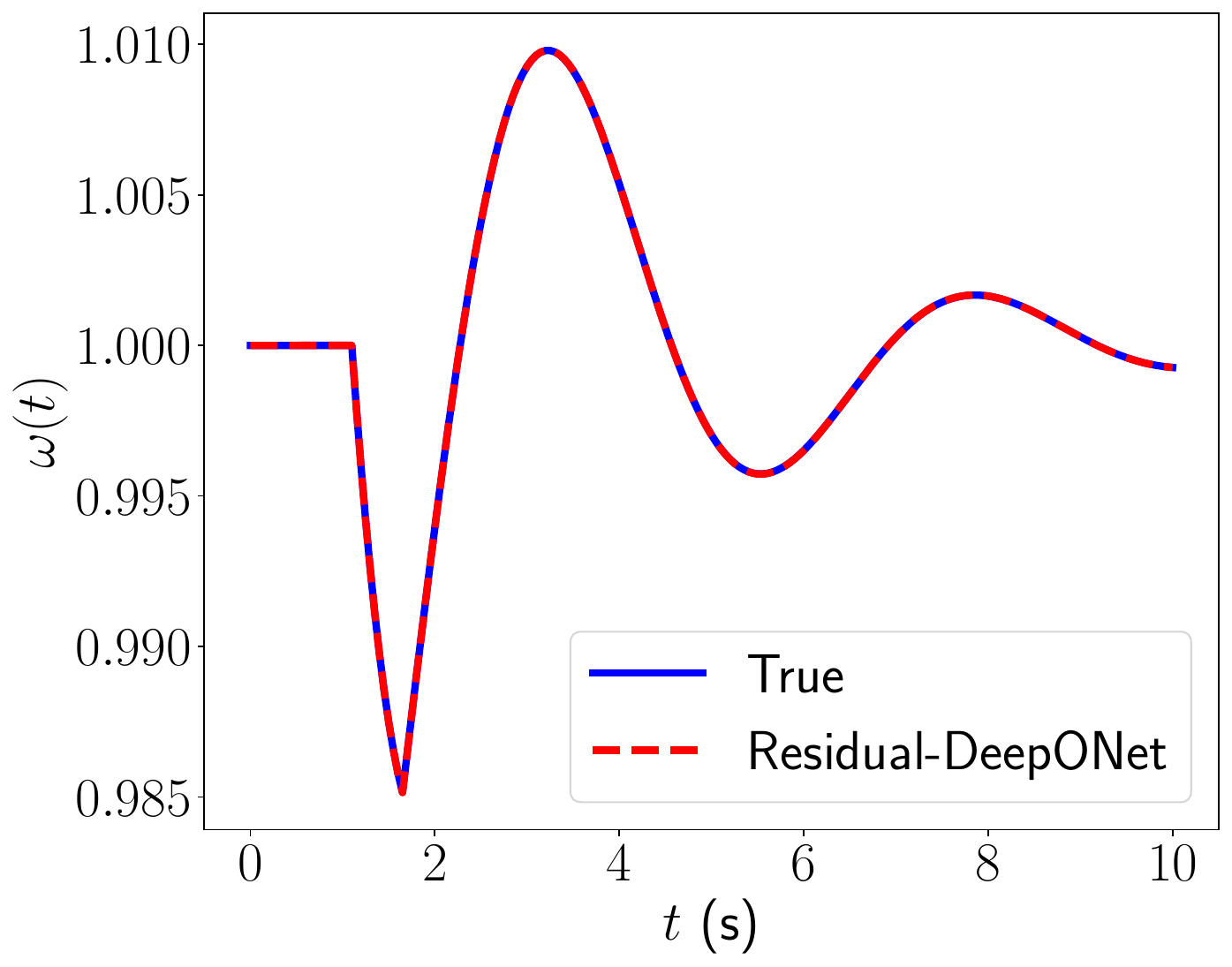}
\end{subfigure}
\begin{subfigure}[b]{0.4\textwidth}
\centering
\includegraphics[width=0.99\textwidth]{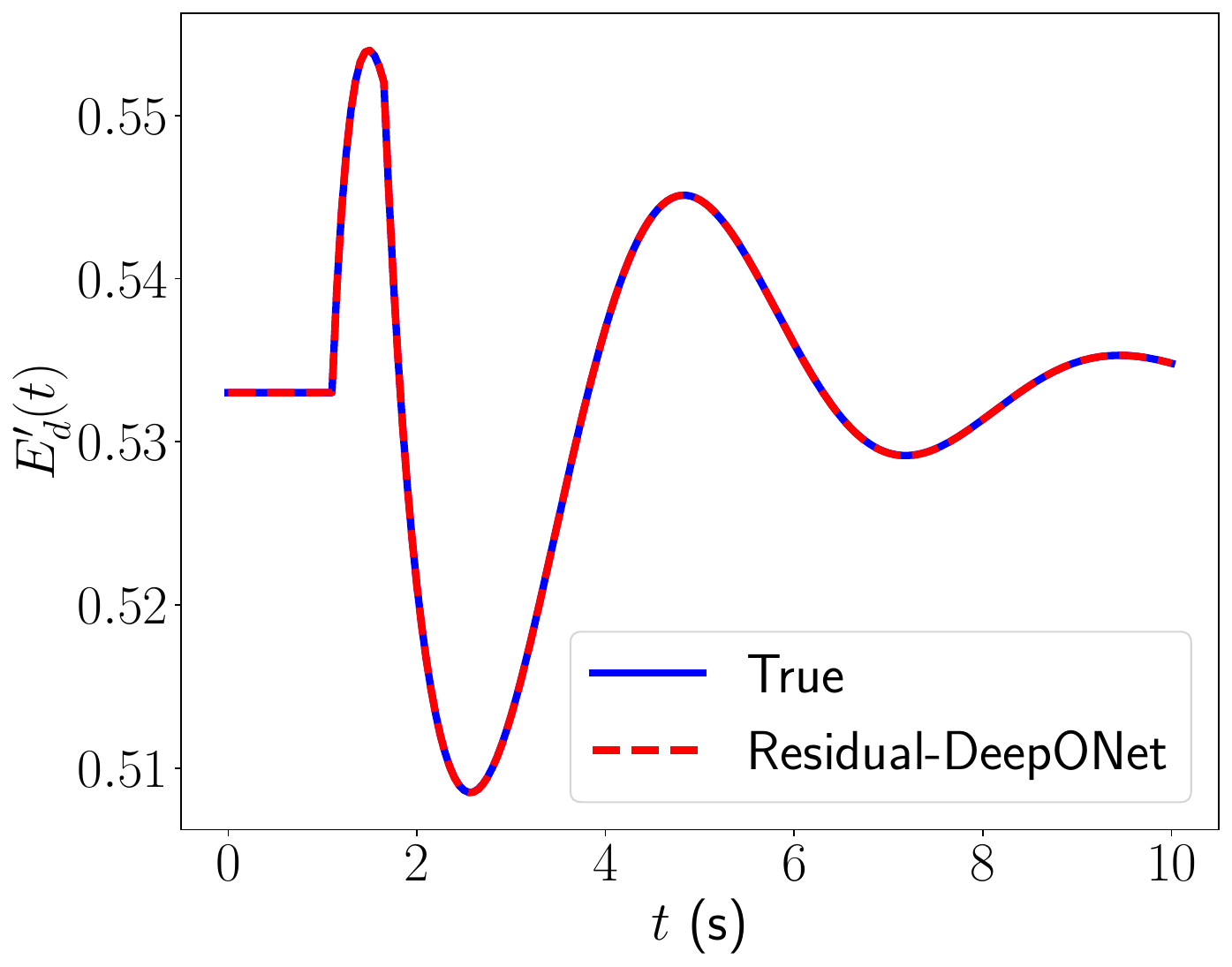}
\end{subfigure}
\caption{Comparison of the residual DeepONet prediction with the true fault trajectory of the SG selected state $(\omega(t), E_d'(t))$ within the irregular partition $\mathcal{P} \subset [0,10]$ (s) of size $200$.}
\label{fig:comparison-states-residual-DeepONet-fault}
\vspace{-1em}
\end{figure}

\begin{figure}[t!]
\centering
\begin{subfigure}[b]{0.4\textwidth}
\centering
\includegraphics[width=0.99\textwidth]{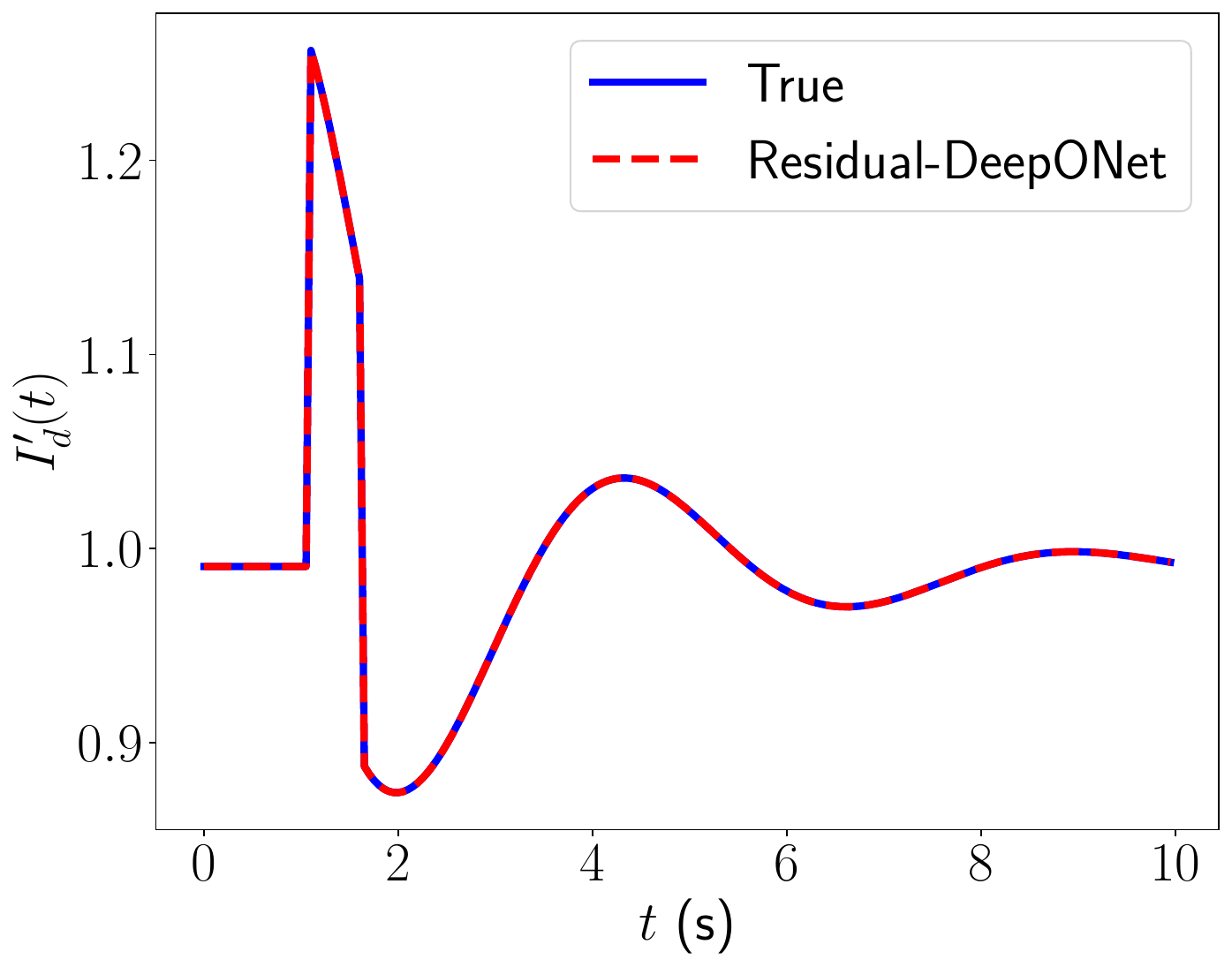}
\end{subfigure}
\begin{subfigure}[b]{0.4\textwidth}
\centering
\includegraphics[width=0.99\textwidth]{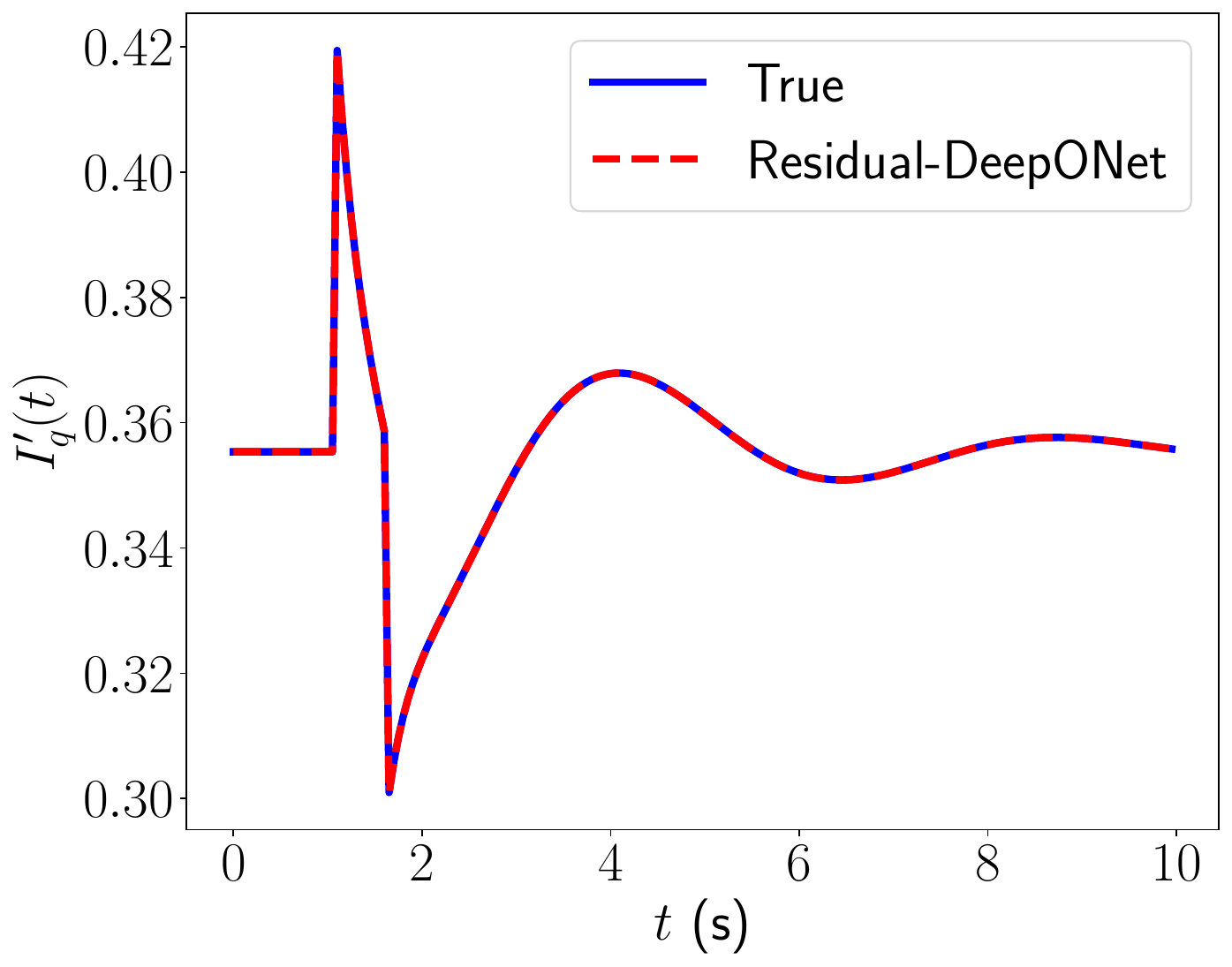}
\end{subfigure}
\caption{Comparison of the resulting network input $y(t) = (I_d'(t), I_q'(t))^\top$ after a random disturbance for the residual DeepONet trained within the irregular partition $\mathcal{P} \subset [0,10]$ (s) of size $200$.}
\label{fig:comparison-inputs-residual-DeepONet-fault}
\vspace{-1em}
\end{figure}

First, we randomly chose a disturbance trajectory from the test dataset and compared the predictions of the proposed residual DeepONet with the actual values. Figs.~\ref{fig:comparison-states-residual-DeepONet-fault} and \ref{fig:comparison-inputs-residual-DeepONet-fault} show the residual DeepONet's prediction of selected state trajectories and network response. The results demonstrate that the proposed framework accurately predicts state trajectories and does not introduce errors when solving for the network equations~\eqref{eq:network-equations} with a disturbance.

To further study the generalization of the residual DeepONet trained with the DAgger algorithm, we computed the mean and standard deviation of the $L_2-$relative error for all $500$ test disturbance trajectories. Table~\ref{table:residual-DeepONet-fault} displays the results for the residual DeepONet trained with DAgger. These results show that the residual DeepONet keeps the mean $L_2-$relative error below $0.01$ \% for all state and network trajectories. This finding suggests that residual DeepONet propagates almost no errors to the network equations~\eqref{eq:network-equations}.

\begin{table}[t]
\centering
\begin{tabular}{ l | c  c  c  c c c}
\hline
& $\delta(t)$ & $\omega(t)$ & $E_d'(t)$ & $E_q'(t)$ & $I_d'(t)$ & $I_q'(t)$\\
\hline

mean $L_2$\% & 8$e$-3\% & 7$e$-4\% & 5$e$-3\% & 3$e$-3\% & 4$e$-3\% & 6$e$-3\% \\
st. dev. $L_2$\% & 4$e$-4\% & 1$e$-4\% & 3$e$-4\% & 2$e$-4\% & 1$e$-4\% & 2$e$-4\% \\
\hline
\end{tabular}
\caption{The mean and standard deviation (st. dev.) of the $L_2$-relative error between the residual DeepONet predicted and the actual response of the (i) synchronous generator state and (ii) network's input for $500$ different disturbance trajectories and within the uniform partition $\mathcal{P} \subset [0,10]$ (s) of constant step size $h = 0.05$.}
\label{table:residual-DeepONet-fault}
\vspace{-0.5em}
\end{table}

\begin{figure}[t!]
\centering
\begin{subfigure}[b]{0.4\textwidth}
\centering
\includegraphics[width=0.99\textwidth]{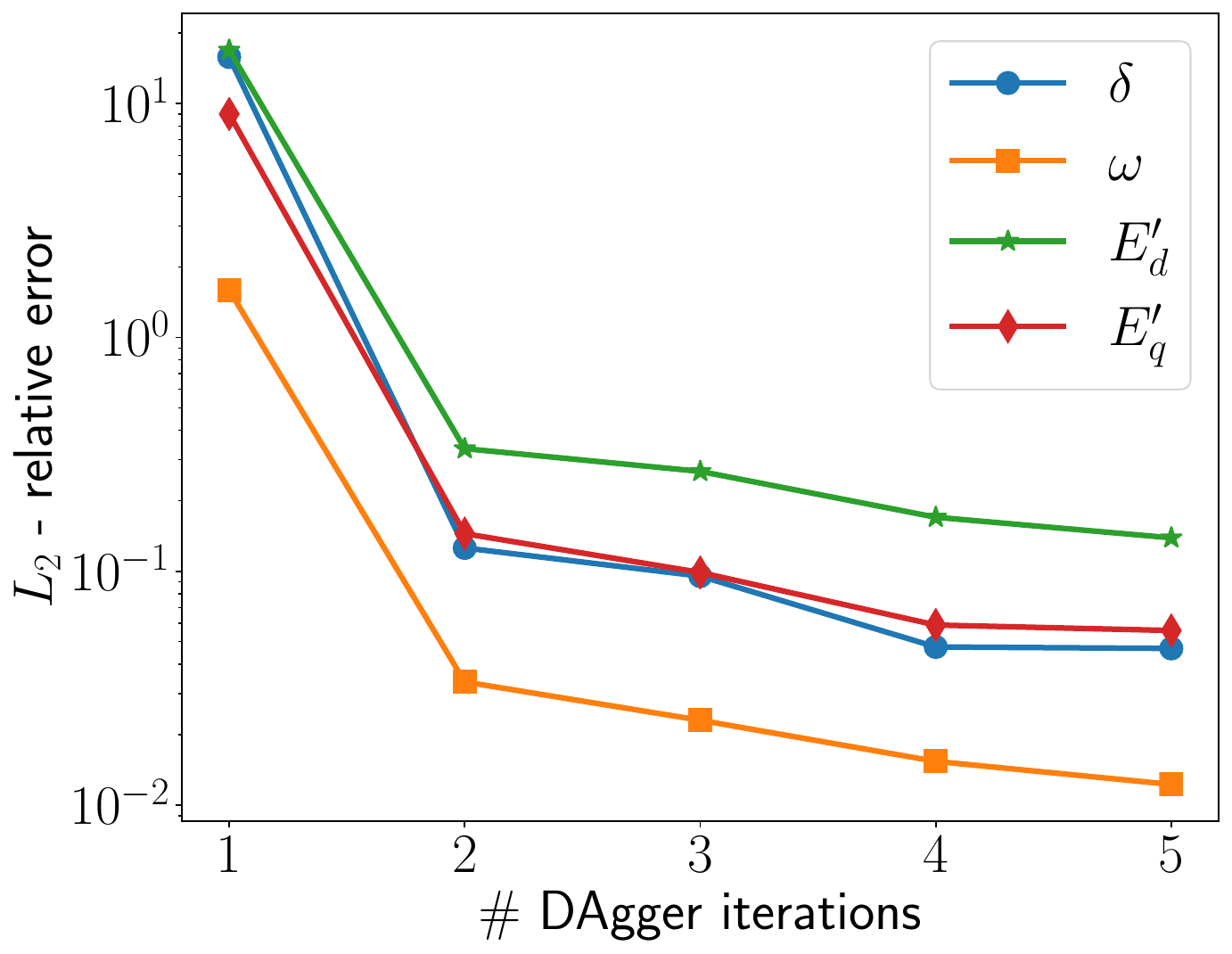}
\end{subfigure}
\begin{subfigure}[b]{0.4\textwidth}
\centering
\includegraphics[width=0.99\textwidth]{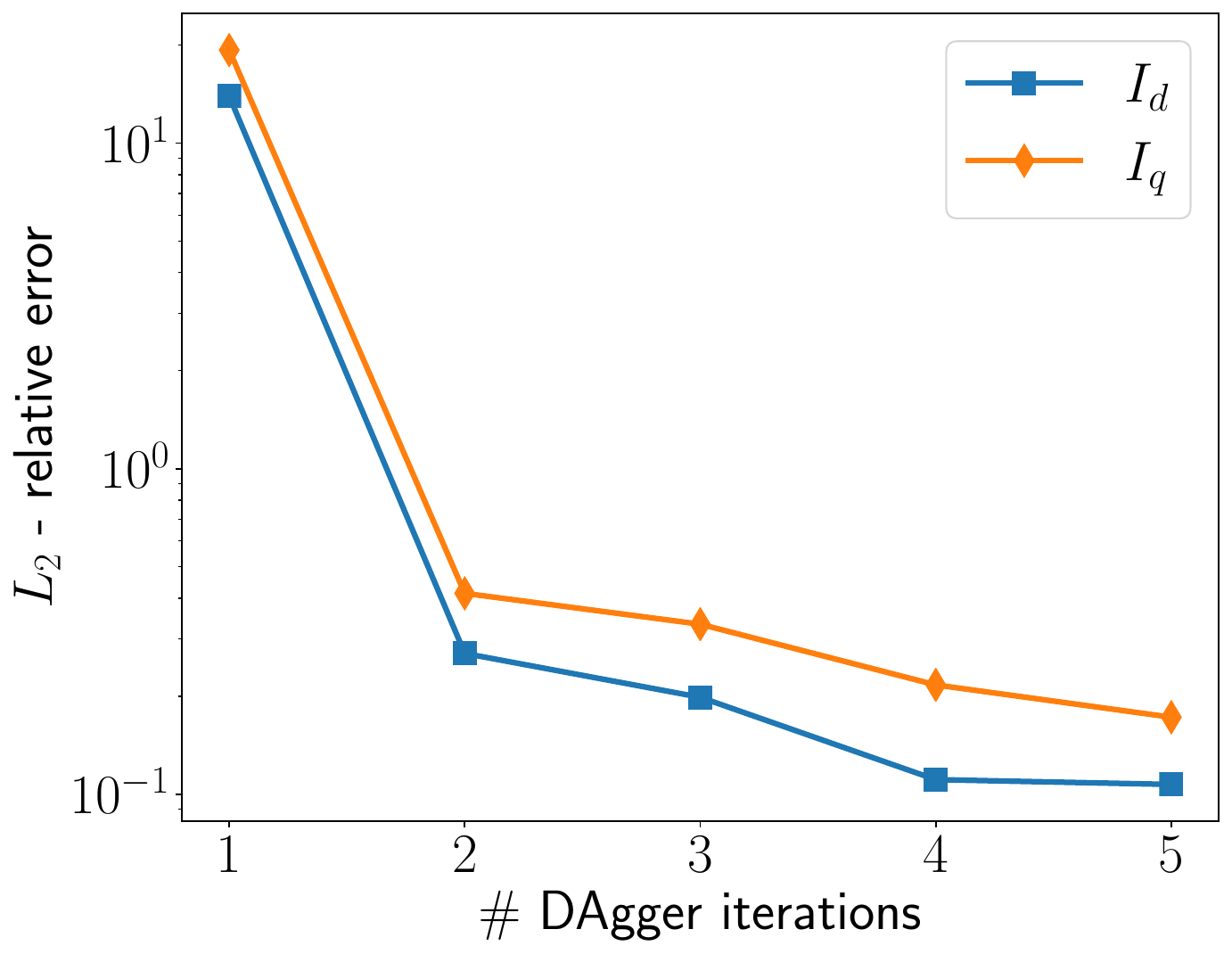}
\end{subfigure}
\caption{The mean $L_2$-relative error~\% of $500$ predicted trajectories with residual DeepONet as a function of the number of DAgger iterations (\textit{left}) generator states $x(t) = (\delta(t), \omega(t), E_d'(t), E_q'(t))^\top$ and (\textit{right}) network inputs $y(t) = (I_d'(t), I_q'(t))^\top$.}
\label{fig:comparison-residual-L2-rel-error-nIters}
\vspace{-1em}
\end{figure}

To conclude this experiment, we studied the effect of the number of DAgger iterations on the mean $L_2-$relative errors for the $500$ test disturbance trajectories. We trained the residual DeepONet with DAgger iterations from the set $\{1, 2, 3, 4, 5\}$. Fig.~\ref{fig:comparison-residual-L2-rel-error-nIters} shows that increasing the number of DAgger iterations decreases the $L_2-$relative errors and error accumulation. This is because the DAgger algorithm aggregates training data samples that the residual DeepONet is likely to encounter when interacting with the infinite bus model.
